%% file: arxiv.tex
\documentclass[runningheads]{llncs}

\usepackage{eccv}

\usepackage{eccvabbrv}

\usepackage{graphicx}
\usepackage{pdfpages}
\usepackage{booktabs}
\usepackage{wrapfig}
\usepackage{amsmath}	
\usepackage{amssymb}
\usepackage{float}
\usepackage{mathtools}
\usepackage{pifont}
\usepackage{multirow}
\usepackage{colortbl}
\usepackage{dsfont} 

\usepackage[accsupp]{axessibility}  

\usepackage{hyperref}

\usepackage{orcidlink}
\input{def}

\begin{document}

\title{\fullname for Zero-shot \\Instance Navigation}

\titlerunning{\fullname for Zero-shot Instance Navigation}

\author{Xinyu Sun\inst{1}\thanks{Equal contribution.} \and
Lizhao Liu\inst{3}$^*$ \and
Hongyan Zhi \and 
Ronghe Qiu\inst{1} \and
Junwei Liang\inst{1,2}\thanks{Corresponding author.}\orcidlink{0000-0003-2219-5569}
}

\authorrunning{X. Sun et al.}

\institute{AI Thrust, The Hong Kong University of Science and Technology (Guangzhou) \and
Department of Computer Science and Engineering, The Hong Kong University of \\Science and Technology \\ \and Tencent AI Lab, Shenzhen, China \\
\email{\{csxinyusun,selizhaoliu\}@gmail.com}, 
\email{junweiliang@hkust-gz.edu.cn}
}

\makeatletter
\renewcommand*{\@fnsymbol}[1]{\ensuremath{\ifcase#1\or *\or \dagger\or \ddagger\or
		\mathsection\or \mathparagraph\or \|\or **\or \dagger\dagger
		\or \ddagger\ddagger \else\@ctrerr\fi}}
\makeatother

\maketitle

\begin{abstract}
  We study zero-shot instance navigation, in which the agent navigates to a specific object without using object annotations for training. 
  Previous \notsure{object navigation approaches} apply the image-goal navigation (\imagenav) task (go to the location of an image) for pretraining, and transfer the agent to achieve object goals using a vision-language model.
  However, these approaches lead to issues of semantic neglect, where the model fails to learn meaningful semantic alignments.
  In this paper, we propose a \fullname (\sexyname) method to improve the semantic understanding ability of navigation agents.
  Specifically, a semantic-enhanced PSL agent is proposed and a \trainlowername strategy is introduced to select goal images that exhibit clear semantic supervision and relax the reward function from strict exact view matching. 
  At inference time, a \inferlowername scheme is designed to preserve the same granularity level of the goal-semantic as training.
  Furthermore, \textit{for the popular HM3D environment}, we present an Instance Navigation (\instancenav) task that requires going to a specific object instance with detailed descriptions, as opposed to the Object Navigation (\objectnav) task where the goal is defined merely by the object category.
  Our PSL agent outperforms the previous state-of-the-art by 66\% on zero-shot \objectnav in terms of success rate and is also superior on the \notsure{new} \instancenav task. Code will be released at \url{https://github.com/XinyuSun/PSL-InstanceNav}.
\end{abstract}

\section{Introduction}
\label{sec:intro}
Visual navigation is a fundamental skill for embodied agents to travel efficiently in complex 3D environments~\cite{imagenav}.
Among different kinds of visual navigation tasks, Zero-shot Object Navigation (ZSON) is a promising task and paves the way for developing a general embodied agent~\cite{anderson2018r2r,wu2018roomnav,thomason2020dialog,gan2022finding,yenamandra2023homerobot,gan2020threedworld,li2023behavior,srivastava2022behavior,li2021igibson}, since it requires zero scene object annotations for training. However, the criteria of the current ZSON task is to evaluate whether the agent can go to the object that matches solely the given category, which remains a considerable gap from \textit{real-life applications} that require the identification of a specific object. 
Therefore, we propose to extend ZSON to Zero-shot Instance Navigation (ZSIN) task in the HM3D environment~\cite{hm3d,hm3dv2}, which requires the agent to go to a unique object instance given detailed text descriptions.
Both the ZSON and ZSIN tasks are very challenging due to the zero-shot constraint, 
in which the category information of goal objects is not feasible during training.

\begin{figure}[t!]
    \centering
    \begin{subfigure}[b]{\linewidth}
        \centering
        \includegraphics[width=\linewidth]{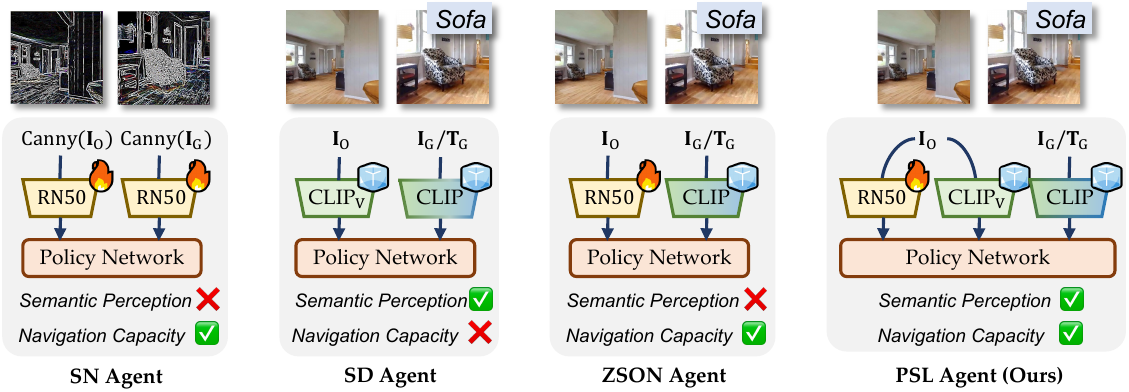}
    \end{subfigure}
    \begin{subfigure}[b]{0.3\linewidth}
        \centering
        \includegraphics[width=\linewidth]{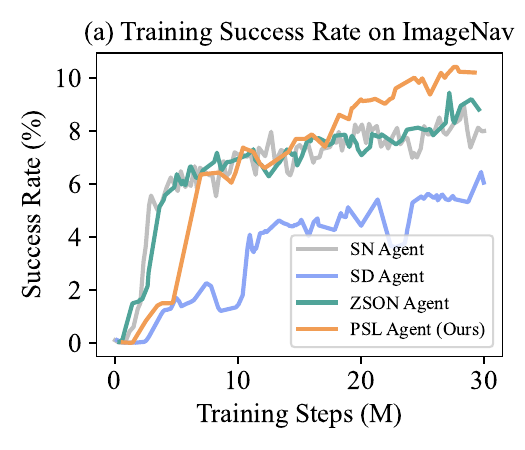}
    \end{subfigure}
    \hfill
    \begin{subfigure}[b]{0.3\linewidth}
        \centering
        \includegraphics[width=\linewidth]{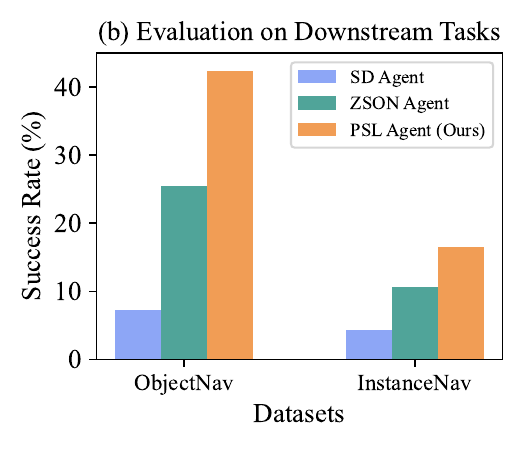}
    \end{subfigure}
    \hfill
    \begin{subfigure}[b]{0.3\linewidth}
        \centering
        \includegraphics[width=\linewidth]{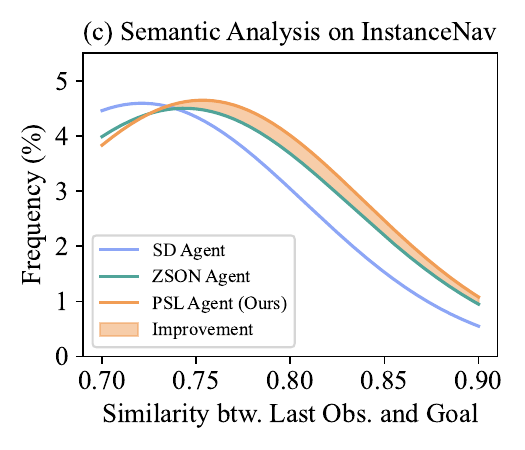}
    \end{subfigure}
    \caption{
    Pilot studies on four navigation agents: Semantic-Non-dominant (SN) agent, Semantic-Dominant (SD) agent, ZSON~\cite{zson} agent, and our \sexyname agent. 
    (a) The agent trained on \imagenav task does not necessarily need to learn the semantic information to obtain a high success rate. 
    (b)-(c) Our \sexyname agent 
    achieves both strong semantic perception and navigation capacity.\protect\footnotemark
    }
    \label{fig:teaser}
\end{figure}

To achieve zero-shot visual navigation, pioneering works~\cite{zer,zson} turn to pre-trained vision-language models such as CLIP~\cite{clip} for its zero-shot capability and propose to train the agent on Image-goal Navigation (\imagenav) pretext task that requires the agent to go to a randomly sampled goal image~\cite{imagenav}. To be specific, using the CLIP vision encoder to obtain a semantic goal from the goal image for training, the ZSON agent~\cite{zson} can be transferred to \objectnav by switching the semantic goal into the CLIP text embedding. 
Although the ZSON agent demonstrates solid navigation capabilities, we argue that it lacks strong semantic perception abilities.

\footnotetext{The SN agent does not support \instancenav task as it lacks a semantic goal encoder.}

To investigate how the agent architecture affects its semantic perception ability and navigation capacity, we conduct pilot studies on four different agent architectures and the results are presented in Fig.~\ref{fig:teaser}. We intriguingly find a Semantic-Non-dominant (SN) agent, with Canny operator\footnote{We consider that the Canny operator has undermined the semantics in the image.} and learnable ResNet50~\cite{he2016deep} encoders to obtain observation and goal embeddings, is able to achieve a competitive success rate on the \imagenav task as the ZSON agent $\left( \text{see Fig.~\ref{fig:teaser} (a)}\right)$. 
In contrast, the Semantic-Dominant (SD) agent with two fixed CLIP vision encoders gets the worst results. Therefore, we conclude that \textit{the \imagenav pre-training task does not necessarily require the agent to perceive semantic information}. In this sense, the ZSON agent with only a learnable observation encoder may possess inferior semantic perception ability, limiting the performance on navigation tasks that heavily rely on finding semantic clues. Therefore, how to effectively improve the semantic perception ability of the navigation agent in the zero-shot setting remains an important but unresolved issue.

In this paper, we propose a \fullname (\sexyname) method to improve the semantic perception and understanding ability of the navigation agent. Our \sexyname approach is comprised of three parts: a \sexyname agent architecture, a \trainlowername strategy, and a \inferlowername scheme.
\textbf{First}, we equip the agent with an additional CLIP vision encoder to \lec{encodes} the semantics in the observation. 
\lec{Subsequently, a \spmfullname (\spmname) is proposed to comprehend the semantic differences between the observation and goal images.}
\textbf{Second}, 
we take advantage of the entropy-minimization technique to select goal images with clear semantic supervision. 
Moreover, we relax the reward function to focus more on semantic correspondence rather than strict geometric matching
\textbf{Third}, considering that the proposed \spmname learns to comprehend the semantic differences of image embeddings during training, we develop a \inferlowername scheme that retrieves image embeddings using text queries. The retrieved image embeddings expand the text embeddings with rich visual priors, facilitating more precise indications.
Without bells and whistles, the proposed \sexyname method achieves state-of-the-art performance on both widely-evaluated \objectnav and our extended \instancenav benchmarks. For example, in the \objectnav task, \sexyname outperforms the ZSON baseline~\cite{zson} and the ESC method~\cite{esc} that uses large-language models by \textbf{16.9\%} and \textbf{3.2\%} success rate, respectively. Furthermore, we verify the proposed \sexyname agent has a stronger semantic perception ability by visualizing the semantic similarities between the agent's last observation and the goal images in Fig.~\ref{fig:teaser}~(c).
\lec{Our main contributions are summarized as follows:}
\begin{itemize}
    \item We investigate the semantic perception ability of different navigation agent architectures and reveal that the commonly-used \imagenav pre-training task does not necessarily require the agent to learn semantic information.
    \item We propose a \lowername method that includes a strong agent architecture, a novel training strategy and an effective inference scheme to improve the agents' semantic perception ability in zero-shot object/instance navigation tasks.
    \item Extensive experiments and sufficient ablation studies on both \objectnav and \instancenav benchmarks demonstrate the superior performance of the proposed \sexyname method over the state-of-the-art.
\end{itemize}

\section{Related Work}

\subsection{Visual Navigation}
Visual navigation is a fundamental and crucial task for robots to perform various embodied tasks~\cite{iGibson, Batra2020rearrange, ai2thorRearrange, BEHAVIOR, BEHAVIOR-1K}. In visual navigation tasks, the agent receives a geometry~\cite{habitat, DDPPO} or semantic~\cite{batra2020objectnav, imagenav} goal and an RGB observation from the camera attached to the agent, and then makes efforts to navigate to this goal according to the observation in each time step. 
These tasks can be divided into different categories, including point-goal navigation~\cite{habitat, DDPPO}, object-goal navigation~\cite{ovrl,ovrlv2,zson,esc,pixelnav,ramrakhya2023pirlnav}, image-goal navigation~\cite{imagenav,ovrl,ovrlv2,zer,crl,nrns,mem-aug,vgm,tsgm,fgprompt}, and vision-language navigation~\cite{R2R, RxR, krantz2020vlnce, krantz2021waypoint, a2nav}, \etc~
Taking image-goal navigation~\cite{fgprompt} as an example, the agent needs to explore the environment, locate the target object specified by the image, and navigate to its proximity.
A common trail for these visual navigation tasks is to train the agent in the simulator~\cite{habitat,habitat2} with photo-realistic environments~\cite{hm3d,hm3dv2,mp3d} and physical engine, and provide the agent with the goal indication as well as its corresponding trajectory annotation.
Since labeling objects in the 3D scenes is labor expensive, recent works~\cite{zer,zson,esc,pixelnav} have explored how to design an object-goal navigation agent without human annotations. These methods are easy to scale up~\cite{zson} and show potential to solve the open set challenge~\cite{ma2023simple} in locating objects.
In this work, we focus on a more challenging task of visual navigation, zero-shot instance navigation, in which the agent is required to navigate to an object specified by detailed descriptions, for example, an instance image~\cite{iin} or text.

\subsection{Vision-Language Model}
Vision-language models (VLM) provide mutual understanding for both vision and language modalities~\cite{udandarao2023visual} and thereby naturally serve as a fundamental bedrock for vision navigation. To build a visual navigation agent, two lines of work are widely employed, namely, representation-based and module-based methods. Representation-based methods~\cite{clip,jia2021scaling,li2021align} aim to co-embed the image and text into a common semantic space. CLIP~\cite{clip} and ALIGN~\cite{jia2021scaling} apply contrastive loss to train the image and text encoders on large-scale pair-wise image text datasets. ALBEF~\cite{li2021align} further proposes to leverage a momentum distillation mechanism that alleviates the weak correlation between noisy image text pairs to improve multimodal semantic learning. As for the module-based methods~\cite{li2019visualbert,li2023blip,zhang2023llama,gao2023llama,liu2023llava,liu2023improved}, they aim to learn a strong multimodal reasoner to simultaneously comprehend vision-language modalities and produce text responses, which are commonly used as the navigation controller~\cite{esc,pixelnav}. Early works such as VisualBERT~\cite{li2019visualbert} apply mask language loss on both image patches and word embeddings to pre-train a multimodal reasoning model. To incorporate the strong reasoning ability of large language models, recent works like BLIP-2~\cite{li2023blip}, LLaMA-Adapter~\cite{zhang2023llama,gao2023llama} and LLaVA~\cite{liu2023llava,liu2023improved} perform visual instruction tuning on pretrained large language models. However, using these module-based vision language models is compute-intensive, making them unsuitable for real-life robotic scenarios that require nearly real-time response. 
\lec{In this paper, we construct the navigation agent using the representation-based VLM and aim to improve the semantic perception ability of the agent.}

\section{Motivation of \fullname}
\label{sec:revisiting}

\imagenav is an important pre-training task to achieve Zero-Shot Instance Navigation (ZSIN) as it provides supervision signals to navigate in 3D environments with visual observation and a semantic indication~\cite{zson}. 
However, we argue that the \imagenav task can be solved without semantic clues, resulting in a sub-optimal pre-training objective.
To verify this, we conduct a preliminary experiment on the HM3D \imagenav dataset that is used to train a zero-shot object navigation agent in previous literature~\cite{zson}. 
We set up four different agents, namely the Layout-Only (LO) agent, Semantic-Only (SO) agent, ZSON agent, and our \sexyname agent.
Among them, the Semantic-Only agent is equipped with two frozen CLIP ResNet-50 encoders to obtain semantic-level observations and goals. 
The Layout-Only agent is adopted with two trainable ResNet-50 encoders and we remove the semantic information in both observation and goal images by using a Sobel operator to calculate gradient images, which leave only layout information like contours and edges (see Fig.~\ref{fig:teaser} for more details).
Results in Fig.~\ref{fig:teaser} (a) show that the agent does not necessarily need to learn semantic information to arrive at an image goal, merely relying on the layout information to perform view matching can obtain a high success rate.

\section{\fullname}
\subsection{Preliminaries}
We follow ZSON~\cite{zson} to adopt image-goal navigation pre-training and perform the zero-shot evaluation on downstream tasks. 
During pre-training, the image-goal navigation (\imagenav) objective provides supervision signals to update the agent. 
Specifically, for each episode with a randomly sampled start point and goal point, the agent is asked to navigate to the goal destination according to a goal image. Empowered by the cross-modal alignment ability of Vision-and-Language models (\ie CLIP~\cite{clip}), we can evaluate the trained agent on the object-goal navigation (\objectnav) dataset by replacing the image embeddings with text embeddings.

\subsection{\fullname Agent Architecture}
We propose a \fullname (\sexyname) agent that emphasizes the reasoning of the semantics differences in the goal and observation.
The key components of \lec{\sexyname} agent include observation and goal encoders, a \spmlowername module and a recurrent policy network.

\begin{figure}[t!]
    \centering
    \includegraphics[width=\linewidth]{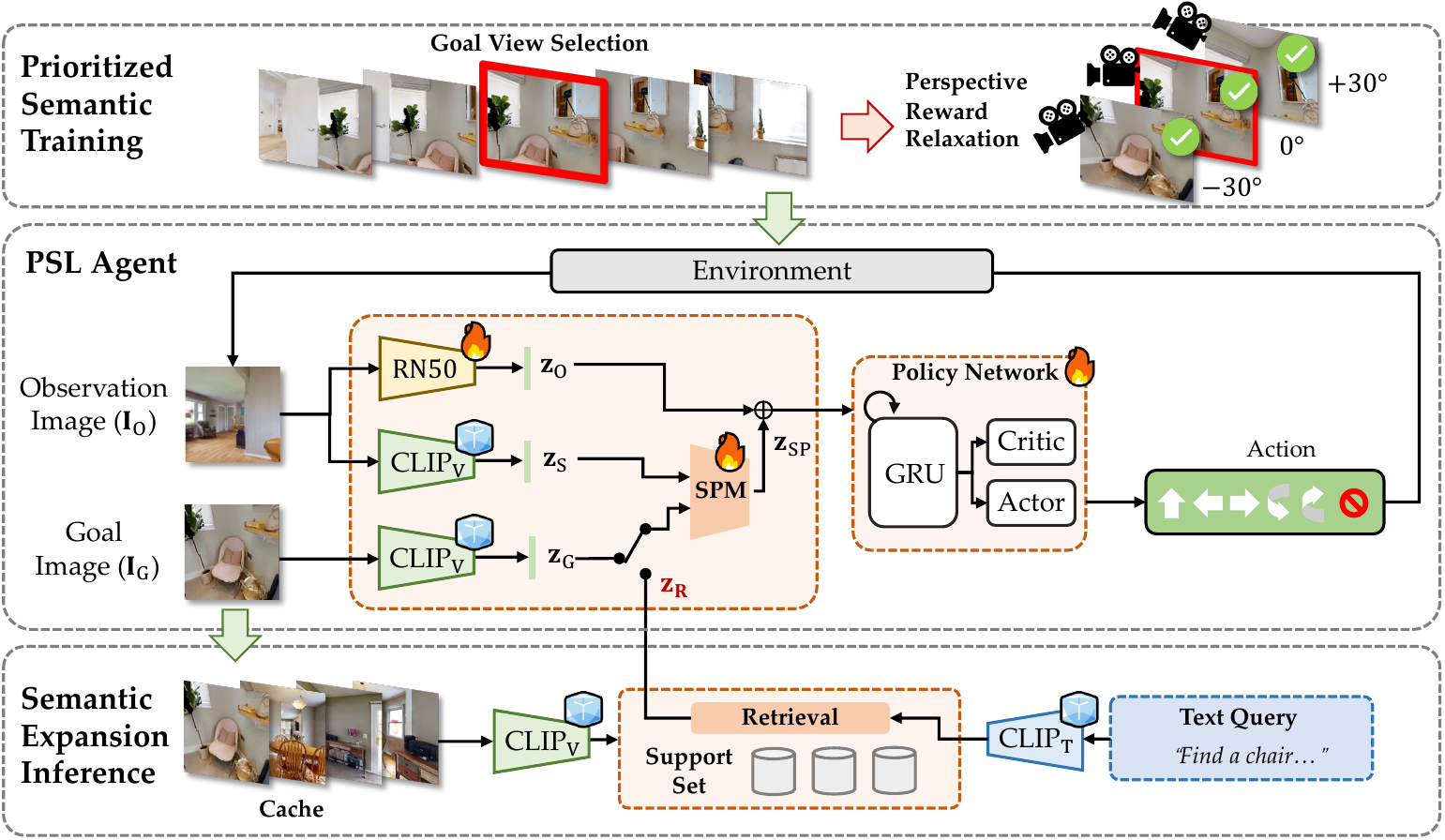}
    \caption{\textbf{Overview of our \sexyname.} 
    1) During \imagenav pre-training, we provide clear semantic supervision signals with the \trainfullname strategy.
    2) Our \sexyname agent exploits the \spmfullname (SPM) to achieve both strong semantic understanding and navigation capacity. 
    3) During inference, a \inferfullname scheme is incorporated to ensure the same semantic granularity of the goal-embedding between training and testing.
    }
    \label{fig:overview}
\end{figure}

\paragraph{\textbf{Observation and Goal Encoders.}}
\lec{Following the ZSON baseline, given the observation image $\bI_\mathrm{O}$ and the goal image $\bI_\mathrm{G}$, we apply a learnable ResNet50 encoder and a fixed CLIP encoder to obtain the observation embedding $\bz_\mathrm{O}$ and the goal embedding $\bz_\mathrm{G}$, respectively.}
\lec{Since with only a trainable ResNet50 as the observation encoder may not effectively learn the semantic information, we additionally leverage a frozen CLIP encoder to extract semantic-level observation $\bz_\mathrm{S}$.}
\lec{Note that the ResNet50 are initialized from the one pre-trained by OVRL~\cite{ovrl}.}

\paragraph{\textbf{\spmfullname.}}
We then introduce the \spmlowername, an MLP layer that \lec{encodes} the semantic correspondence between $\bz_\mathrm{G}$ and $\bz_\mathrm{S}$ into a low-dimension feature. Specifically, it takes both $\bz_\mathrm{G}\in \mathbb{R}^{C_1}$ and $\bz_\mathrm{S}\in \mathbb{R}^{C_1}$ as input and produce \lec{semantic perception embedding} $\bz_\mathrm{SP}\in \mathbb{R}^{C_2}$, where $C_2 < 2\times C_1$. The \spmlowername reduces 
\lec{the feature dimension,}
serving as a bottleneck to condense the critical semantic \lec{correspondence} in both the goal image and observation image.

\paragraph{\textbf{Policy Network.}}
Based on the \lec{semantic perception embedding} $\bz_\mathrm{SP}$ 
\lec{and the observation embedding $\bz_\mathrm{O}$}, we train a navigation policy $\pi_\theta$ using reinforcement learning (RL):
\begin{equation}
    \bs_t, \bh_t = \pi_{\theta}(\bz_\mathrm{SP}\oplus \bz_\mathrm{O} \oplus \ba_{t-1}|\bh_{t-1})
\end{equation}
where $\bs_t$ is the embedding of the agent's current state, which is then used to predict an action distribution among six action categories, including \texttt{MOVE\_FORWARD}, \texttt{TURN\_LEFT}, \texttt{TURN\_RIGHT}, \texttt{STOP}, \texttt{LOOK\_UP} and \texttt{LOOK\_DOWN}. $\bh_{t-1}$ denotes hidden state of the recurrent layers in policy $\pi_\theta$ from the previous step.
Following previous methods~\cite{zer, zson}, we adopt an actor-critic network to predict state value $c_t$ and action $a_t$ using $s_t$ and train it end-to-end using PPO~\cite{PPO}. 
More details can be found in the Appendix.

\begin{figure}[t!]
    \centering
    \includegraphics[width=\linewidth]{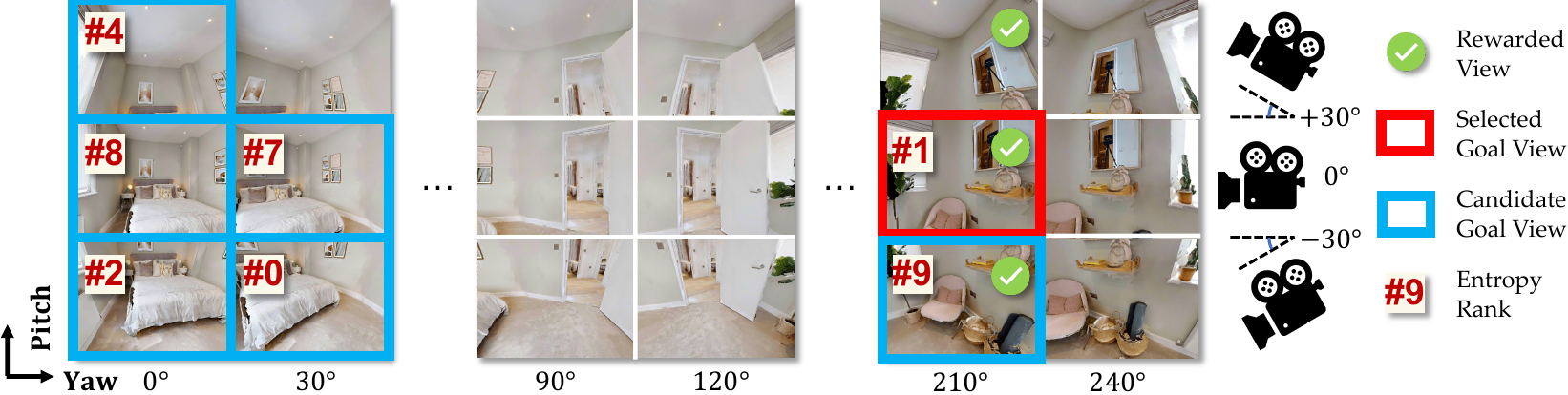}
    \caption{Illustration of the perspective relaxation and goal view selection approach in our training strategy. 
    The goal view image set is expanded with additional yaw and tilt views using perspective relaxation. 
    Top-10 views with minimum entropy are circled in red with their ranking.}
    \label{fig:data-views}
\end{figure}

\subsection{Prioritized Semantic Training Strategy}
Previous literature~\cite{iin,modiin} has pointed out that a large portion of image goals in the \imagenav task lead to ambiguity (e.g. looking at walls).
We provide statistical analysis on the distribution of \imagenav goal images to support this speculation (see Fig. II in Appendix). In particular, We select 6 object categories that widely exist in HM3D scenes, as well as two additional categories ``wall'' and ``room'' to represent images with ambiguous semantic content. 
Then we \lec{use CLIP to} perform zero-shot classification on all goal images by selecting the category with maximum image-text similarity. As presented in previous literature, in the original HM3D \imagenav dataset, most of the goal images cannot be categorized as common object categories; instead, they are classified as meaningless indications (\ie ``wall'' and ``room''). 
We believe that the unreasonable distribution of goal images aggravates the semantic neglect issue.

\paragraph{\textbf{Entropy-minimized Goal View Selection.}}

To alleviate the semantic neglect issue, we propose an entropy-based goal view selection method to ensure a dominant object exists in the goal images.
Specifically, we randomly select a base orientation at the goal point and rotate the agent for $\Omega$ times uniformly to render images in different view angles. 
Then we obtain image embeddings $\{\bv_\omega \mid \omega \in \Omega\}$ using CLIP~\cite{clip} vision encoder and calculate the similarity between image embeddings with all text embeddings of the candidate object categories. We follow the standard setting of Object-Goal navigation approaches~\cite{ovrl,ovrlv2,esc} to use 6 common objects $\mC$ in the 3D simulator and compute query vectors $\{\bq_{c} \mid c \in \mC\}$ using CLIP text embedding.
The objective is to select a goal view $\omega^{*}$ that minimizes the entropy of probability distribution among all object categories:

\begin{equation}
    \omega^{*} = \underset{\omega \in \Omega}{{\arg\min}} -\frac{1}{\log(| \mC |)} \sum_{c\in \mC} \bp_c{\log\bp_c},
\label{eq:entropy}
\end{equation}
where $\bp_c=\mathrm{softmax}\left(g(\bv_w, \bq_c)\right)$ is the class probability \wrt $c^\text{th}$ object category, $g(\ba, \bb) = \tau \cdot \frac{\ba^T\bb}{\lVert \ba \rVert_2 \lVert \bb \rVert_2}$ is the scaled cosine similarity function and $\tau$ is the temperature.
As shown in Fig. II in the Appendix, the goal images with ambiguous semantics are significantly reduced.
Additionally, image distribution among all object categories becomes more balanced.

\paragraph{\textbf{Perspective Reward Relaxation.}}
Choosing goal views with rich semantic content enables the agent to focus on meaningful regions in the goal images. However, the \imagenav reward~\cite{zer,zson} still motivates the agent to strive for an exact match between the goal image and the observation image. 
We tackle this challenge by introducing a perspective reward relaxation technique that cancels exact matching. 
Specifically, we augment the goal view selection process by rendering additional circles of images around the agent at different pitch angles.
As illustrated in Fig.~\ref{fig:data-views}, we then perform goal view selection among all rendered views that vary in pitch and yaw angles. 
Furthermore, we rewrite the \textbf{reward function} of PPO training by relaxing the perspective matching requirement:

\begin{align}
\begin{split}
R^\text{\sexyname}_t = &\underbrace{\gamma^{suc} * \mathds{1}\{d_t < \epsilon^d \}}_{\textit{reach the goal location or not}} + \underbrace{\gamma^{suc} * \mathds{1}\{d_t < \epsilon^d \} * \mathds{1}\{(\mathrm{extract}_\mathbf{Y}(\ba_t) < \epsilon^a \}}_{\textit{match the goal view or not}} \\ & + \underbrace{ r_d(d_t,d_{t-1})+ \mathds{1}\{d_t < \epsilon^d \}*\mathrm{extract}_\mathbf{Y}(r_a(\ba_t,\ba_{t-1})) }_{\textit{closer to the goal or not}} - \gamma^{delay},
\end{split}
\label{eqn:reward}
\end{align}
which is comprised of four parts. \textbf{First}, a success reward $\gamma^{suc}=5$ is given when distance to goal metric $d_t$ is smaller then threshold $\epsilon^d$. \textbf{Second}, an angle success reward is given when the agent reaches the goal location and faces the goal orientation, where $\mathrm{extract}_\mathbf{Y}(\ba_t)$ means the component of quaternion $\ba_t$ rotating around the $\mathbf{Y}$ axis. We only encourage the agent to head to the target in the x-z plane while ignoring its pitch heading. \textbf{Third}, a dense reward compose of reduced distance $r_d$ and angle $r_a$ to the target at time step $t$. \textbf{Last}, a delay penalty $\gamma^{delay}$ is to encourage efficiency.

\subsection{Semantic Expansion Inference Scheme}
In the inference stage, the agent receives a series of text descriptions of the target object. Existing methods directly replace the image-goal embedding with the text-goal embedding of the input text description to perform zero-shot evaluation. 
However, the granularity gap~\cite{liang2022mind} between image and text embeddings could considerably impair the evaluation performance~\cite{li2023decap}.
Since it is non-trivial to obtain text annotations for all goal images (28.8M) during training, 
we opt to enrich the text goal with fine-grained visual features to reduce the gap between image-goal pre-training and text-goal evaluation.
We name this method \inferlowername.

\textit{During training}, we maintain a support set $\mV=\{\bv\in\mathbb{R}^d \mid \bv_i \neq \bv_j, g(\bv_i, \bv_j) < \lambda \}$ without using any category information. As we expect the selected embeddings to be as dissimilar as possible from the ones already in the set, we set a threshold $\lambda=0.8$ to filter representative image embeddings, resulting in around 0.1M vectors. These embeddings approximate the distribution of goal images, and they encode rich information from a variety of object instances, regarding their detailed attributes (\eg color, shape, texture, and material).
\textit{During inference}, given a language description of the target object, we leverage the CLIP text encoder to produce a feature $\bz_\mathrm{T}$. Then we query a goal embedding using the retrieval operation:
\begin{equation}
    \bz_\mathrm{R} = \sum_{\bv_i \in \mV} \frac{\mathrm{exp}(g(\bz_\mathrm{T}, \bv_i))}{\sum_{\bv_j \in \mV}\mathrm{exp}(g(\bz_\mathrm{T}, \bv_j))} \ast \bv_i.
\label{equation:weighted-sum}
\end{equation}
We perform a weighted sum based on the similarity score to avoid inferior performance produced by the nearest neighbor querying~\cite{li2023decap}.

\section{Experiments}
In this section, we first compare the proposed \sexyname agent with previous zero-shot methods in the \objectnav task that target finding arbitrary object instances of the same category,
and then investigate how it performs when instructed to navigate to a specific object in the \instancenav task. 
We also conduct extensive ablation experiments to study the effectiveness of different modules in our method empirically.

\subsection{Experimental Setup}
\paragraph{\textbf{Datasets.}}
During pre-training, we utilize the 7,200,000 episodes generated by ZSON~\cite{zson} from HM3D~\cite{hm3dv2}. We randomly select 4 goal images for each episode from 10 candidates with minimal entropy in Eqn.~\ref{eq:entropy}. For \objectnav evaluation, we adopt the habitat2022 challenge dataset with 2000 episodes and 6 object categories. For \instancenav, we first evaluate the agent on IIN dataset~\cite{iin}, an image-goal track with 795 unique instances in 1000 test episodes, and then extend it to the text-goal setting. The agent is tested on 20 scenes excluded from the 800 training scenes.

\paragraph{\textbf{Agent Configuration.}}
We follow the configuration in~\cite{zson} with an agent height of 0.88m, radius of 0.18m, and a single $640\times480$ RGB sensor with a 79$^\circ$ horizontal field-of-view (HFOV) placed 0.88m from the ground. For \instancenav, in the image-goal setting, the agent receives a goal image captured in a random camera configuration that does not match the agent's own. In the text-goal setting, the agent receives text description of both intrinsic and extrinsic attributes (\eg \textit{``Find a double bed and its material is a beige bamboo frame. There are only two objects around the bed: a table and a window."}).
For \objectnav, the agent receives the category text as instruction like ``\textit{Find a bed}''.

\paragraph{\textbf{Evaluation Metrics.}}
We evaluate the agent's performance by two metrics: Success Rate (SR) and Success Rate Weighted by Path Length (SPL). An episode is marked as a success if the agent navigates to the goal object and executes the \texttt{STOP} action within 0.1m to any viewpoints of the goal object. The SR metric is the average number of success indicators among all episodes. The SPL metric is defined as $\frac{1}{N}\sum_{i=1}^N\textrm{SR}_i\frac{l_i^*}{\mathrm{max}(l_i,l_i^*)}$, where $l_i$ denotes to the agent's path length and $l_i^*$ denotes to the ground-truth path length. We report the average number over three runs with different random seeds.

\subsection{Experiments on the \objectnav Task}
In Table~\ref{tab:sota-objectnav}, we report the zero-shot evaluation results on the HM3D \objectnav task. 
Our \sexyname agent outperforms the ZSON baseline~\cite{zson} by $16.9\%$ in SR ($42.4\%$ \vs $25.5\%$), thanks to our training strategy and inference scheme design. 
More importantly, our method exceeds the LLM-based methods in the first time. Compared to the PixelNav method~\cite{pixelnav} that learns to navigate to a point deduced by a powerful multi-modal LLM~\cite{gpt4}, our \sexyname agent achieves $+4.5\%$ improvement in SR. Furthermore, taking into account mapping-based methods that perform LLM-guided frontier-based exploration~\cite{yamauchi1997frontier}, we see $+6.0\%$ improvements against L3MVN~\cite{yu2023l3mvn} and $+3.2\%$ against ESC~\cite{esc}. Our method shows great potential to be applied in the real environment as it does not require accurate GPS coordinates to build a map. Besides, our LLM-free architecture is more efficient.

\begin{table}[h]
\centering
\footnotesize
\caption{Comparison with state-of-the-art methods on the ObjectNav task. 
Our \sexyname surpasses both LLM-based and Mapping-based methods in terms of Success Rate (SR).}
\label{tab:sota-objectnav}
\setlength\tabcolsep{3pt}
\begin{tabular}{@{}lcccccc@{}}
\toprule
Method                      & with Mapping & with LLM & LLM     & Extra Sensors & SR            & SPL           \\ \midrule
L3MVN~\cite{yu2023l3mvn}    & \with        & \with    & GPT-2   & Depth, GPS     & 35.2          & 16.5          \\
PixelNav~\cite{pixelnav}    & \without     & \with    & GPT-4   & -            & 37.9          & 20.5          \\ 
ESC~\cite{esc}              & \with        & \with    & GPT-3.5 & Depth, GPS     & 39.2          & 22.3          \\ \midrule
CoW~\cite{cow}              & \with        & \without & -       & Depth, GPS     & 6.1           & 3.9           \\
ProcTHOR~\cite{procthor}    & \with        & \without & -       & Depth, GPS     & 13.2          & 7.7           \\
ZSON~\cite{zson}            & \without     & \without & -       & -             & 25.5          & 12.6          \\
\sexyname(Ours)             & \without     & \without & -       & -             & \textbf{42.4} & \textbf{19.2} \\ \bottomrule
\end{tabular}
\end{table}

\subsection{Experiments on the \instancenav Task}
\begin{figure}[t!]
    \centering
    \includegraphics[width=0.85\linewidth]{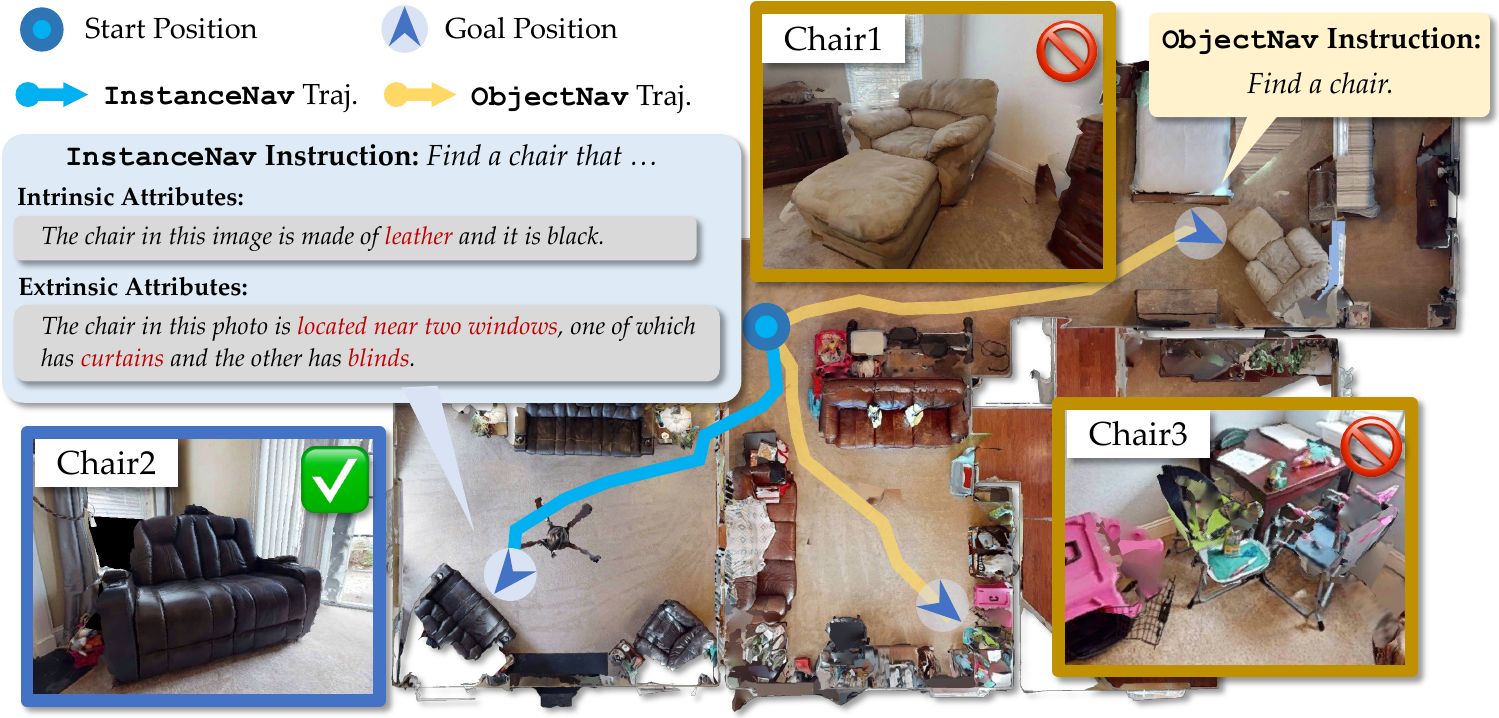}
    \caption{Comparison of the \instancenav task with the \objectnav task. The \objectnav task allows the agent to navigate to any chair in the scene, while the \instancenav task has only one target object specified by detailed attribute descriptions.}
    \label{fig:instancenav}
\end{figure}

The \objectnav task assigns multiple available destinations for each episode.
It is not sure whether the agent can distinguish object instances.
In comparison, we extend the zero-shot object navigation task to the zero-shot instance navigation in the \textit{widely-used HM3D environment},
as illustrated in Fig.~\ref{fig:instancenav}.
In this task, the agent receives a textual description of the targeted instance or an instance image as the goal. It needs to understand the detailed attributes of an object instance and navigate to an exact location. 

\paragraph{\textbf{Dataset Preparation.}}
Previous works on the instance-level object navigation task focus on discrete graph-based environments~\cite{zhu2021soon} and are limited by complicated human annotation process~\cite{cow}. 
In this work, instead, we utilize the recent success of the Generative Vision-and-Language Models (GVLMs)~\cite{gpt4,llava,cogvlm} to automatically build the open-vocabulary text-goal setting for the \instancenav task on the popular HM3D environment. 
Specifically, we randomly select a goal view for each validation episode to render an image of the goal object in the Instance Image Navigation (IIN) dataset~\cite{iin}. Each episode in the IIN dataset corresponds to a unique goal object instance. 
To specify each object instance, we follow a previous attempt~\cite{majumdar2023findthis} to separate text descriptions into two aspects: Intrinsic Attributes and Extrinsic Attributes. Intrinsic attributes cover inherent characteristics of the object, such as shape, color, and material. Extrinsic attributes describe the environment surrounding the object, which is used to determine instances with similar intrinsic attributes. 
We instruct a GVLM model (\ie CogVLM~\cite{cogvlm}) to generate both types of attributes according to the instance image with a hand-crafted prompt (See Sec. A in the Appendix for details).
The original ground truth trajectories are preserved to construct the additional \instancenav test set.
In total, the test set of the text-goal setting comprises 1,000 episodes featuring 795 unique objects across 36 scenes.

\paragraph{\textbf{Results under the Text-goal Setting.}} 
To demonstrate our \sexyname agent's ability to understand complicated text descriptions and navigate to specified instances,
we compare it with state-of-the-art zero-shot object-goal navigation methods under the text-goal setting of the \instancenav task. 
In Table~\ref{tab:sota-instance-objectnav}, we find that most of the existing zero-shot object navigation methods struggle in this task, including both map-based methods CoW~\cite{cow} ($1.8\%$ in SR) and LLM-based method ESC~\cite{esc} ($6.5\%$).
All these methods perform inferior to the methods that receive semantic goals (\ie ours and ZSON), indicating the effectiveness of end-to-end approaches that are pre-trained on large-scale unlabeled data in this task.
Similar to our approach, we also transfer a well-trained ImageNav method (OVRL~\cite{ovrl}) to perform the text-goal evaluation using our \inferlowername scheme. We find that though good at the \imagenav task, the OVRL model performs poorly when given retrieved image goals, indicating that the learnable goal encoder and observation encoder failed to encode semantic-level information for the \instancenav task.
In comparison, our \sexyname further brings a significant improvement over the ZSON baseline ($+5.9\%$ in SR).

\begin{table}[]
    \centering
    \footnotesize
    \caption{Comparison with state-of-the-art methods in the text-goal track of the \instancenav task. 
    We report the baseline results based on the released code and models. 
    $^\dagger$We perform evaluation with our proposed \inferlowername scheme.}
    \label{tab:sota-instance-objectnav}
    \setlength\tabcolsep{3pt}
    \begin{tabular}{@{}lcccccc@{}}
    \toprule
    Method & Backbone & with LLM & with Mapping  & Extra Sensors & SR   & SPL \\ \midrule
    CoW~\cite{cow}              & ViT-Base  & \without & \with    & Depth, GPS & 1.8  & 1.1 \\
    GoW~\cite{esc}              & ViT-Base  & \without & \with    & Depth, GPS & 7.2  & 4.2    \\
    ESC~\cite{esc}              & ViT-Base  & \with    & \with    & Depth, GPS & 6.5  & 3.7    \\
    OVRL$^\dagger$~\cite{ovrl}  & ResNet-50 & \without & \without & -          & 3.7  & 1.8 \\
    ZSON~\cite{zson}            & ResNet-50 & \without & \without & -          & 10.6 & 4.9 \\
    \sexyname(Ours)             & ResNet-50 & \without & \without & -          & \textbf{16.5} & \textbf{7.5} \\ \bottomrule
    \end{tabular}
\end{table}

\label{sec:sota}
\paragraph{\textbf{Results under the Image-goal Setting.}}
We then compare our \sexyname agent with other methods under the original Image-goal setting\cite{iin}. The agent is tasked to navigate to an object instance specified by an image.
This task is different from the \imagenav task since the goal images are shot from perspectives with different camera configurations, including different pitch angles and HFOV.
Existing state-of-the-art methods for this task heavily rely on supervision signals from scene object annotations and labeled viewpoints. 
For a fair comparison, we mainly focus on the performance of zero-shot approaches.
Results in Table~\ref{tab:sota-iin} show that our method brings significant improvements on state-of-the-art image-goal navigation methods ($+13.1\%$ compared to FGPrompt~\cite{fgprompt} and $+8.4\%$ compared to ZSON~\cite{zson}). 
Meanwhile, we significantly reduce the gap with the supervised methods like OVRL-V2 in terms of SPL ($11.4\%$ \vs $11.8\%$), even though we use a much smaller vision backbone (ResNet-50 \vs ViT-Base).

\begin{table}[h]
    \centering
    \footnotesize
    \caption{Comparison with state-of-the-art methods in the image-goal track of the \instancenav task. In this track, the ``Supervised'' mark means human labels on the objects are used. $^\dagger$We re-implement OVRL based on released pre-trained weight.}
    \label{tab:sota-iin}
    \setlength\tabcolsep{3pt}
    \begin{tabular}{@{}llcccc@{}}
    \toprule
    Method   & Backbone  & Supervised & Pre-training Data & SR   & SPL  \\ \midrule
    RL Agent~\cite{iin}      & ResNet-18 & \with      & -        & 8.3  & 3.5  \\
    OVRL-V2~\cite{ovrlv2}    & ViT-Base  & \with      & Gibson   & 24.8 & 11.8 \\ \midrule
    OVRL-V2~\cite{ovrlv2}    & ViT-Base  & \without   & Gibson   & 0.6  & 0.2  \\
    OVRL$^\dagger$~\cite{ovrl}         & ResNet-50 & \without   & HM3D     & 8.0  & 4.2  \\
    FGPrompt~\cite{fgprompt} & ResNet-9  & \without   & HM3D     & 9.9  & 2.8  \\
    ZSON~\cite{zson}         & ResNet-50 & \without   & HM3D     & 14.6 & 7.3  \\
    \sexyname(Ours)          & ResNet-50 & \without   & HM3D     & \textbf{23.0} & \textbf{11.4} \\ \bottomrule
    \end{tabular}
\end{table}

\subsection{Ablation Study}
\paragraph{\textbf{Effectiveness of \fullname.}}
We first conduct ablation studies under the image-goal setting of the \instancenav task. 
We start from the ZSON~\cite{zson} baseline, with a fixed CLIP goal encoder and a learnable ResNet-50 observation encoder. Then we apply the entropy-minimized goal view selection algorithm to select meaningful goal images during pre-training data pre-processing. In Table~\ref{tab:ablation-LSD}, this modification brings $+2.1\%$ improvement in success rate. We then add the perspective reward relaxation technique, which enhances goal orientation with an alternative pitch angle. 
This approach aims to prevent the agent from being immersed in exactly matching the goal view and observations, neglecting the semantic correlation between them.
However, we discovered that with the vanilla model design, the augmented image goal confuses the agent and decreases the navigation success rate (from 12.7\% to 10.8\%). 
We argue that this phenomenon is due to the inferior agent design that neglects to reason semantic differences in the goal images and observation images.
Incorporating with the \sexyname agent alleviates this issue. We found that it boosts the navigation success rate by $+3.8\%$ with the goal view selection approach, but more impressively, it cooperates well with the perspective relaxation mechanism and contributes a lot to the improvement ($+9.3\%$ in SR and $+4.2\%$ in SPL). 

\begin{table}[h]
\centering
\footnotesize
\caption{Ablation studies of different components in our \fullname (\sexyname) agent and \trainfullname (PST) strategy under the Image-Goal setting of the \instancenav task. The default entry is marked in \colorbox{mygray}{\color{black}gray}. ``SPM'': \spmfullname; ``GVS'': Goal View Selection; ``PRR'': Perspective Reward Selection.}
\label{tab:ablation-LSD}
\setlength\tabcolsep{4pt}
\begin{tabular}{@{}cccccccccc@{}}
\toprule
\multicolumn{1}{l}{\multirow{2}{*}{}}                                          & PSL & \multicolumn{2}{c}{PST} & \multicolumn{2}{c}{ZSIN-image} & \multicolumn{2}{c}{ZSIN-text} & \multicolumn{2}{c}{ZSON} \\ \cmidrule(l){2-10} 
\multicolumn{1}{l}{}                                                           & SPM & GVS & PRR & SR & SPL & SR & SPL & SR & SPL \\ \midrule
\textbf{ZSON}                                                                  & \without & \without & \without & 12.7 & 6.5 & 10.6 & 6.5 & 25.5 & 12.6 \\ \midrule
\multirow{4}{*}{\textbf{\begin{tabular}[c]{@{}c@{}}PSL\\ (Ours)\end{tabular}}} & \with    & \without & \without & 19.5 & 7.9 & 13.0 & 5.6 & 33.7 & 15.8 \\
                                                                               & \without & \with    & \without & 14.8 & 7.7 & 11.8 & 6.1 & 30.4 & 14.7 \\
                                                                               & \with    & \with    & \without & 16.5 & 6.5 & 12.3 & 5.7 & 35.0 & 18.1 \\
                                                                               & \with    & \with    & \with    & \textbf{22.0} & \textbf{10.7} & \textbf{16.5} & \textbf{7.5} & \textbf{42.4} & \textbf{19.2} \\ \bottomrule
\end{tabular}
\end{table}

\paragraph{\textbf{Ablation on \inferfullname.}}
To alleviate the granularity inconsistency issue of text and image embeddings during testing, we propose a \inferlowername scheme for our \sexyname agent.
We perform elaborate ablation studies under the text-goal setting to investigate its effectiveness. We compare two different implementations of support set in the \inferlowername scheme, including collecting goal images from the image-goal navigation dataset, which refers to the ImageNav support set; and from the instance image navigation dataset, namely the IIN support set.
In Table~\ref{tab:support-set}, we start with ablating the detailed configurations of the retrieval module and have two main findings: 
1) using the support set images from the limited categories of objects as the dataset defined yields inferior results ($11.1\%$ in SR). Instead, the diversity of the ImageNav support set brings $+1.3\%$ improvements.
2) directly replacing the image features with text queries performs poorly ($6.6\%$ in SR) due to the modality gap~\cite{liang2022mind}. Retrieving image features to augment the text queries yields better goal embeddings that share the same level of semantic granularity as training, thus significantly improving the navigation success rate by $+9.9\%$.

\begin{table}[h]
\centering
\footnotesize
\caption{Ablation studies of the \inferfullname (SEI) scheme under the Text-Goal setting of the \instancenav task.}
\label{tab:support-set}
\setlength\tabcolsep{3pt}
\begin{tabular}{@{}cccccc@{}}
\toprule
 \sexyname Agent & SEI Scheme & Support Set & \#Supp. Vec. & SR   & SPL \\ \midrule
\without & \with     & IIN         &     3.5K     & 11.1 & 5.4 \\
\without & \with     & ImageNav    &     0.1M     & 12.4 & 6.6 \\
\with    & \without  & ImageNav    &     0.1M     & 6.6  & 2.7 \\
\cellcolor{mygray}\with & \cellcolor{mygray}\with & \cellcolor{mygray}ImageNav & \cellcolor{mygray}0.1M & \cellcolor{mygray}\textbf{16.5} & \cellcolor{mygray}\textbf{7.5} \\ \bottomrule
\end{tabular}
\end{table}

\section{Conclusion}
We present \sexyname, a zero-shot approach for instance navigation. \sexyname is built upon the open-vocabulary and cross-modal ability of CLIP, and targets addressing the limited semantic perceiving ability of existing methods.
Our method is composed of three different parts, including a \sexyname agent with a \spmlowername module to reason the semantic differences between the observation and goal, a \trainlowername strategy with clear semantic supervision to train the \sexyname agent, and a \inferlowername scheme to maintain the granularity of the goal-semantic.
We investigate how this agent acts when instructed to navigate to a specific object by extending the zero-shot object navigation task to a zero-shot instance navigation task. 
We find that most existing methods struggle in this task, while our \sexyname agent outperforms other approaches.

\section*{Acknowledgements}
This work was supported by the Meituan Academy of Robotics Shenzhen, the Guangzhou Municipal Science and Technology Project (No. 2024A03J0619) and the National Natural Science Foundation of China (No. 62306257).

\bibliographystyle{splncs04}
\bibliography{main}



\section*{Supplementary material}

# Prioritized Semantic Learning for Zero-shot Instance Navigation (Supplementary Materials)

Xinyu Sun[1*], Lizhao Liu[3*], Hongyan Zhi, Ronghe Qiu[1], and Junwei Liang[1,2†]

[1] AI Thrust, The Hong Kong University of Science and Technology (Guangzhou)
[2] Department of Computer Science and Engineering, The Hong Kong University of Science and Technology
[3] Tencent AI Lab, Shenzhen, China
{csxinyusun,selizhaoliu}@gmail.com, junweiliang@hkust-gz.edu.cn

We organize our supplementary materials as follows:

- In Section A, we demonstrate the detailed collection process and statistics of the InstanceNav benchmark.
- In Section B, we provide more results on the pilot study.
- In Section C, we provide additional ablation studies on our proposed semantic expansion inference scheme.
- In Section D, we ablate different selections of the Vision-and-Language Model.
- In Section E, we perform additional ablation on the semantic perception module in our PSL agent.
- In Section G, we provide more implementation details of our PSL agent.
- In Section H, we provide more visualization results.

## A  Details and Statistics of HM3D InstanceNav Benchmark

In this section, we visualize the statistics of our proposed InstanceNav benchmark. In order to construct the language instructions for the text-goal settings, we randomly select a rendered goal image for each goal instance provided by the IIN dataset [3], resulting in 795 images. Since the goal images are ensured to capture perfect coverage of the instance, we are able to generate attribute descriptions for each image using a multi-modal large language model [10]. For instance, we instruct the LLM to caption each image using the prompt "*Describe the material and color of the {object} in this image. Describe the surrounding objects of the {object} in this image.*". The {object} placeholder is replaced by the specific object category of the instance. In Figure I, we visualize the word frequency in our generated text descriptions.

For a fair comparison with our proposed method, we re-implement existing state-of-the-art ObjectNav methods and evaluate them on the InstanceNav benchmark. We leverage the official code to implement the evaluation protocol for the CoW [2] and ESC [13] baseline. We instruct the agent to find the object

---

[*] Equal contribution.
[†] Corresponding author.

category as they can not process complex language input. For the OVRL [11] baseline, as the authors haven't released the trained `ObjectNav` model on the HM3D dataset, we adopt to train an `ImageNav` agent based on the released self-supervised ResNet-50 vision encoder, then transfer it to the `InstanceNav` by retrieving image goals with our semantic expansion inference scheme.

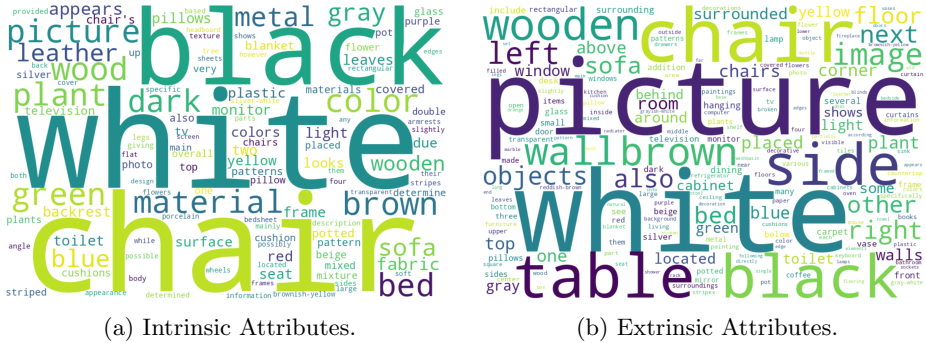

(a) Intrinsic Attributes.                       (b) Extrinsic Attributes.

Fig. I: Word cloud visualization of the attribute descriptions in the Text-Goal settings of the `InstanceNav` task.

## B  More Results on the Pilot Studies

As mentioned in the main paper, we conduct statistical analysis on the distribution of `ImageNav` goal images (see Fig. II). In particular, We select 6 object categories that widely exist in HM3D scenes, as well as two additional categories "wall" and "room" to represent images with ambiguous semantic content. Then we use CLIP to perform zero-shot classification on all goal images by selecting the category with maximum image-text similarity. In Fig. II, in the original HM3D `ImageNav` dataset, most of the goal images cannot be categorized as common object categories; instead, they are classified as meaningless indications (*i.e.,* "wall" and "room"). We believe that the unreasonable distribution aggravates the semantic neglect issue.

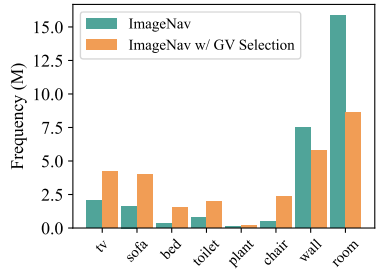

Fig. II: Unreasonable category distribution of goal images in the `ImageNav` task. Our proposed Goal View (GV) Selection releases this issue.

We further provide more results on the pilot studies. To further inspect the navigation capacity and semantic perception ability of the agent, We apply the EigenCAM [6] for visualization of different observation encoders. The results are

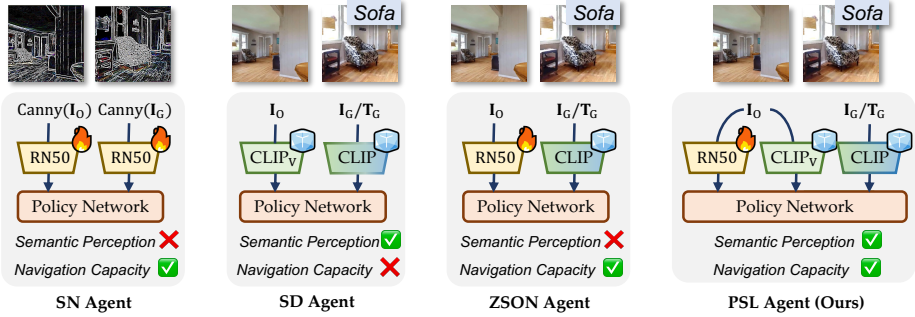

Fig. III: Architecture overview of four different navigation agents: Semantic-Non-Dominant (SN) agent, Semantic-Dominant (SD) agent, ZSON [5] agent, and our PSL agent.

shown in Fig. IV. We also put different agent architectures in Fig. III for quick reference. We have the following observations: **First**, for the learnable observation encoders RN50 of all agents, we identify that they focus on the contours and edges information, showing that the geometric cues are important for strong navigation capacity. **Second**, for the fixed observation encoders $CLIP_V$ from the semantic-only agent and our PSL agent, we find that they pay more attention to the object semantics, mostly on the central part, verifying that it is reasonable to incorporate the semantic observation encoder to improve the semantic perception ability.

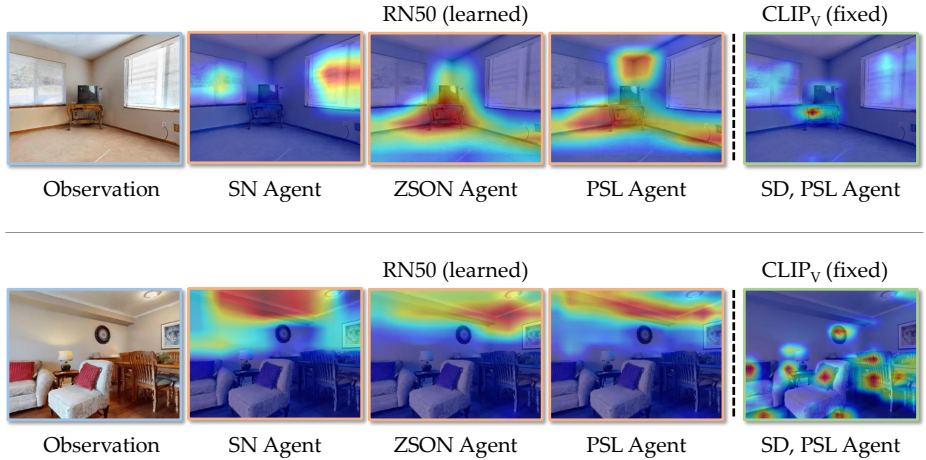

Fig. IV: EigenCAM visualization for the observation encoder in four different agents.

## C  Ablation Studies on the Semantic Expansion Inference Scheme

***Comparison with Baseline Methods.*** We first compare our semantic expansion inference scheme with other techniques that also refine the text input using external models, including:

- *Parsed Attribute Bindings*: this baseline approach leverages the NLTK model [4] to parse the nouns and their corresponding attribute bindings in the descriptions. After that, all bindings are organized in order to form the instruction.
- *Synthesised Image*: a state-of-the-art language-guided image generation method is incorporated to generate an image $\mathbf{I}_G^*$ based on the language description. Specifically, we report the results using the Stable-Diffusion v1.5 model [9] and a compositional generation model [8]. Then, we use the CLIP vision encoder to extract image features $\mathbf{z}_G^*$ and enhance the original text embeddings $\mathbf{z}_T$ by performing a weighted sum $\lambda_1 \mathbf{z}_G^* + \lambda_2 \mathbf{z}_T$.
- *Retrieved Feature*: our proposed method. A goal embedding $\mathbf{z}_R$ is retrieved from the image embeddings in the support set $\mathcal{S}$ using the text embedding as the query, as discussed in the previous section.

In Table I, we report the results of different baseline methods and our proposed retrieval-based semantic expansion inference approach. We observe a marginal decrease in performance for the synthesized image goals, despite the SynGen model exhibiting greater sensitivity to detailed object attributes, as shown in Figure V. We attribute this phenomenon to the inconsistent background introduced by the image generation model. Moreover, the synthesized images can not provide the agent with structural priors in the specific domain. In contrast, our method alleviates this issue by introducing a support set collected during unsupervised pre-training. The images in such a support set are more realistic, facilitating rich visual priors for the navigation task.

| Goal | Ext. Model | SR | SPL |
|---|---|---|---|
| Category Text | - | 8.2 | 4.3 |
| Parsed Attribute Bindings | NLTK [4] | 10.8 | 5.1 |
| Synthesized Image | SD-v1.5 [9] | 9.0 | 4.2 |
| Synthesized Image | SynGen [8] | 10.0 | 4.5 |
| Retrieved Feature (Ours) | NLTK [4] | **12.4** | **6.6** |

Table I: Comparison of different text goal refinement techniques for inference.

***Effect of the Support Set Size.*** We provide an additional ablation study to investigate the effect of the support set size in our semantic expansion inference scheme. During the training phase, we store all goal image embeddings, resulting

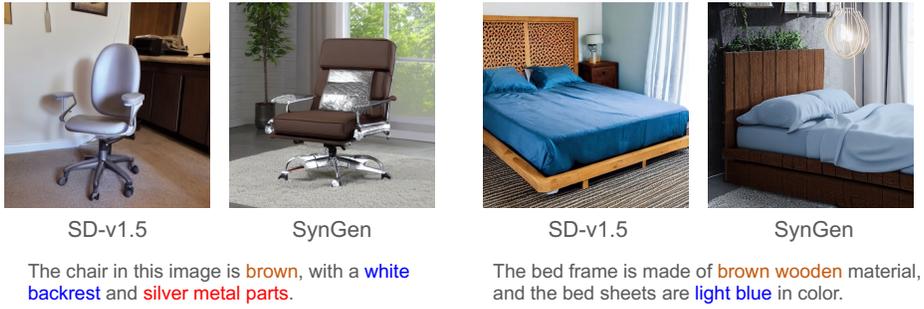

SD-v1.5  SynGen  SD-v1.5  SynGen

The chair in this image is brown, with a white backrest and silver metal parts.

The bed frame is made of brown wooden material, and the bed sheets are light blue in color.

Fig. V: Visualization of synthesized images using SD-v1.5 and SynGen.

in a large support set with 28,800,000 support embeddings. From these, we randomly select four subsets of varying sizes: 1,000 (1K), 10,000 (0.01M), 100,000 (0.1M), and 1,000,000 (1M) support embeddings, respectively. We then perform the semantic expansion inference scheme to retrieve expanded goal embeddings with these variants of support set on the Text-Goal setting of `InstanceNav` task and on the `ObjectNav` task. Results in Figure VI reveal that the size of the support set plays a crucial role by enhancing the diversity of the support embeddings, thereby incorporating new perspectives and a broader array of object images. However, beyond a certain threshold, specifically at 0.1M embeddings, the incremental number of embeddings ceases to improve navigation performance. The extra embeddings tend to introduce redundancy and noise, adversely affecting performance rather than enhancing it.

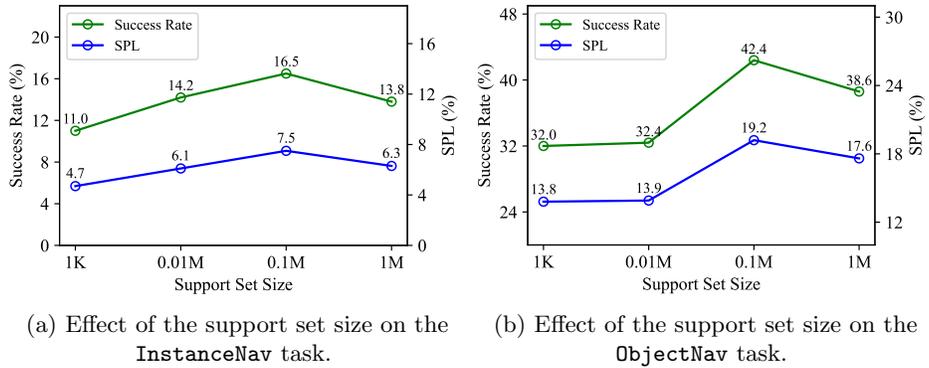

(a) Effect of the support set size on the `InstanceNav` task.

(b) Effect of the support set size on the `ObjectNav` task.

Fig. VI: Ablation study on the support set size on two different tasks.

## D  Ablation Study on the Vision-and-Language Model Selection

In this section, we perform ablation on different Vision-and-Language Models (VLM) that connect `ImageNav` with `InstanceNav`. We adopt to replace the VLM with the SigLIP [12] model introduced by previous work [1]. We then compare this substitution variant with the origin ZSON baseline on the `InstanceNav` task. Results in Table II show that changing the VLM model to an improved version contributes a marginal improvement in SR, but leads to a $-0.7\%$ drop in SPL. For a fair comparison with ZSON, we opt to keep CLIP in our pipeline.

| VLM | SR | SPL |
|---|---|---|
| SigLIP [12] | **11.1** | 4.2 |
| CLIP [7] | 10.6 | **4.9** |

Table II: Effect of the selection of VLM model.

## E  Ablation Studies on the Semantic Perception Module

We provide an additional ablation on the semantic perception module to investigate the effect of the output feature dimension. Travel back to the semantic perception module, it takes in both goal embedding $\mathbf{z}_\mathrm{G} \in \mathbb{R}^{C_1}$ and semantic embedding $\mathbf{z_S} \in \mathbb{R}^{C_1}$ from observation to produce a semantic perception embedding $\mathbf{z}_\mathrm{SP} \in \mathbb{R}^{C_2}$, where $C_2 < 2 \times C_1$. In practice, we set $C_1 = 1024$ which is consistent with CLIP embedding, and we compare the `InstanceNav` results with different dimension numbers $C_2 \in [256, 1024, 2048]$. In Table III, we find that keeping $C_1$ lower than $C_2$ yields better navigation results. We believe this module serves as a bottleneck to condense useful semantic perception results.

| Feature Dim. | SR | SPL |
|---|---|---|
| 256 | 15.8 | 7.1 |
| 1024 | **16.5** | **7.5** |
| 2048 | 14.4 | 6.9 |

Table III: Results on different output feature dimension in the semantic perception module.

## F  More Evaluation Results

In this section, We provide additional evaluation results on object categories unused in goal view selection. We test the agents using a subset of the MP3D `ObjectNav` dataset that **excludes** 6 object categories used for selecting goal views during training, resulting in episodes with 15 **different** object categories. In Table IV, our method still significantly outperforms the ZSON baseline.

We also provide evaluation results on the vanilla `ImageNav` tasks. The vanilla `ImageNav` task requires the agent to go to a location specified by a random image, while the ImageGoal version of the ZSIN task requires the agent to navigate to an object specified by its corresponding image. Substantial improvements in Table V demonstrate the effectiveness of our PSL agent in the `ImageNav` task.

| Method     | SR              | SPL            |
|------------|-----------------|----------------|
| ZSON       | 7.6             | 3.6            |
| PSL (Ours) | **18.9** (+11.3) | **6.4** (+2.8) |

Table IV: `ObjectNav` results.

| Method     | SR              | SPL             |
|------------|-----------------|-----------------|
| ZSON       | 26.9            | 21.7            |
| PSL (Ours) | **32.5** (+5.6) | **23.1** (+1.4) |

Table V: `ImageNav` results.

## G  More Implementation Details

***Training and Evaluation.*** We provide the implementation details of our PSL agent in this section. The agent is trained for 1G steps following ZSON [5] on the `ImageNav` task, and save checkpoint per 10M steps. We evaluate all checkpoints in each downstream task and select the best model. The reported value is an average number over 3 runs with different seeds. All experiments are conducted on 16 Nvidia RTX-3090 GPUs with 16 environments per GPU for training and on 1 GPU with 10 environments for evaluation.

***The Semantic Expansion Inference Scheme.*** During the evaluation on the `ObjectNav` task and the Text-Goal setting of the `InstanceNav` task, we instruct the PSL agent to go to a destination using a retrieved goal embedding with our semantic expansion inference scheme. For the `ObjectNav` task, given a navigation instruction "*Find a {object}*", we extract the text feature $\mathbf{z}_T$ of the category text "*{object}*" using the CLIP text encoder. Then, we directly retrieve a goal embedding using the text embedding $\mathbf{z}_T$. For the Text-Goal setting of the `InstanceNav` task, we leverage the NLTK [4] model to parse the intrinsic attribute descriptions and extract bindings. The query text embedding is the averaged combination of CLIP text features of intrinsic bindings $\mathbf{z}_T^{\text{int}}$ and extrinsic attribute descriptions $\mathbf{z}_T^{\text{ext}}$, for instance, $\mathbf{z}_T = (\mathbf{z}_T^{\text{int}} + \mathbf{z}_T^{\text{ext}})/2$.

***Visualization of semantic expansion inference.*** In Fig. VII, we provide t-SNE visualization of the embeddings used in our semantic expansion inference scheme. We present a triplet example where we retrieve an embedding from the support set using a text query. The retrieved embedding shows less difference from the ground-truth goal image embedding compared to the original text query. Our support set encompasses the object categories in `InstanceNav` and extends to a broader range of embeddings from various perspectives, contributing to the performance gain in navigation success rate.

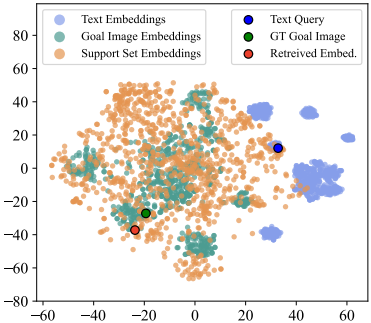

Fig. VII: t-SNE vis. of the embeddings used by semantic expansion inference scheme.

### G.1  Qualitative Analysis

***Qualitative examples for PSL agent.*** In Fig. VIII we present qualitative examples of our PSL agent navigating to two different object instances given detailed language instruction (*e.g.,* , "*Find the plant made of the black pot and grayish-white branches*, around with a transparent glass dining table and four white chairs."). The agent navigates across rooms, determining relative object instances according to the instruction, and finally stops near the object. We find that given detailed language instruction, the agent is able to differentiate between objects of the same category. For instance, in the first case, the agent bypasses a plant with "*a white bottle and pink flower*" which differs from the intrinsic attribute description. After a left turn near the first plant, the agent arrives at its destination and faces the correct plant instance with "*black pot and grayish-white branches*". In another case, the agent hovers for a while in the intersection of bedroom and dining room to find the "*white chair*" and last the episode at "*several white chairs with a glass dining table*" that mentioned in the extrinsic attributes.

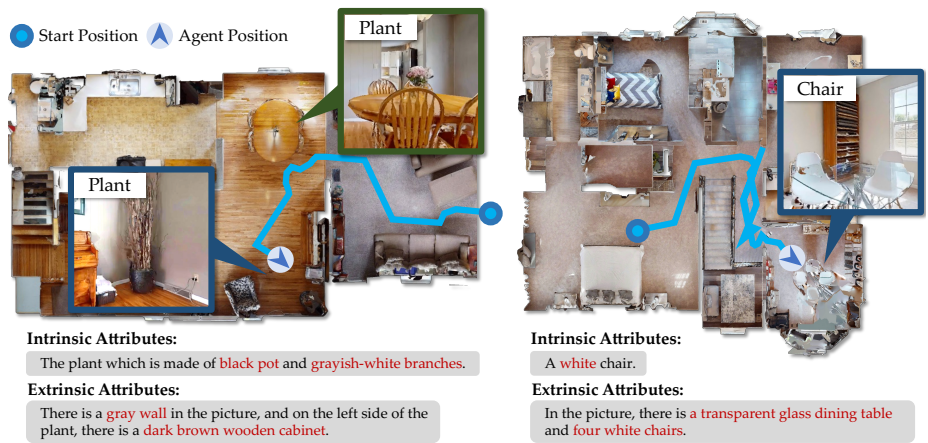

Fig. VIII: **Qualitative examples** for our PSL agent navigating to an object instance according to intrinsic and extrinsic object attributes in the `InstanceNav` dataset. For each trial, the agent is initialized at a random position in the room and given language instruction "*Find the ...*".

## H    More Qualitative Results

In this section, we provide more qualitative results on the `InstanceNav` task.

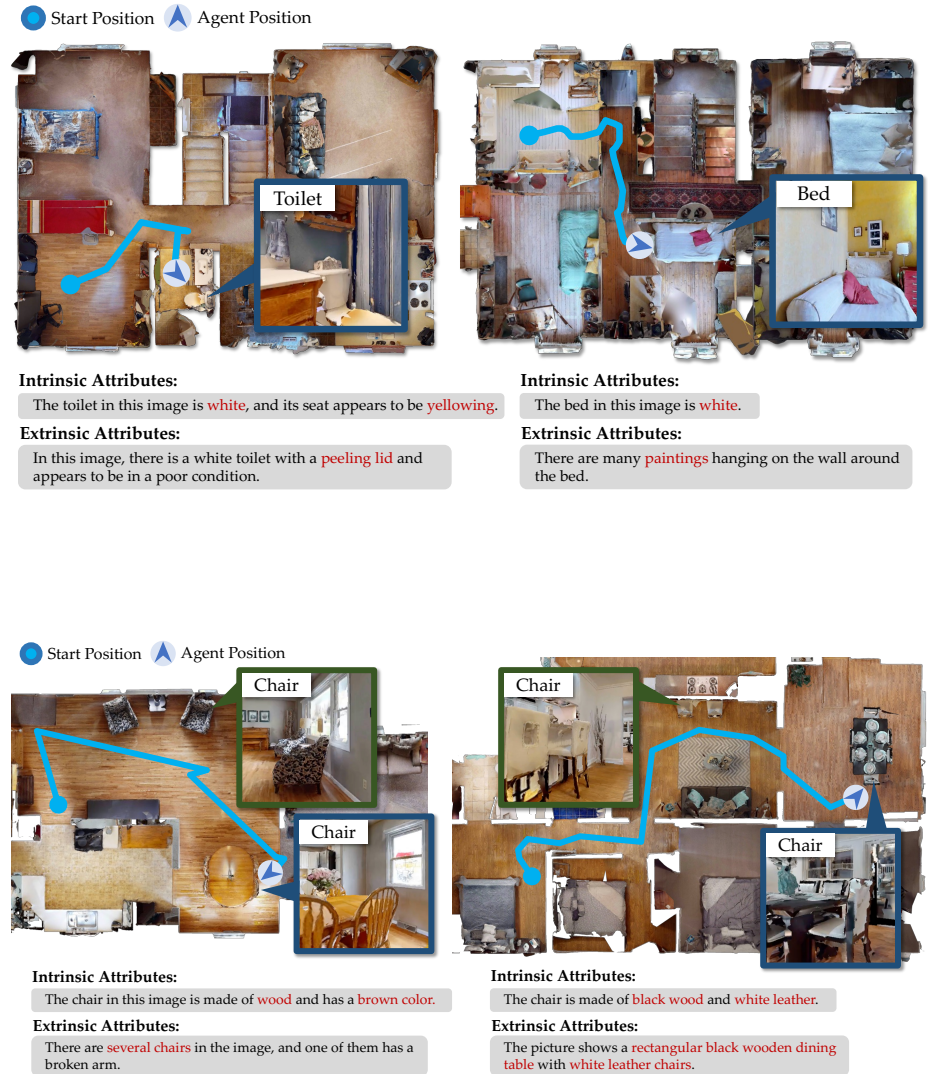

Fig. IX: Addition qualitative results on the `InstanceNav` task.

## References

1. Ehsani, K., Gupta, T., Hendrix, R., Salvador, J., Weihs, L., Zeng, K.H., Singh, K.P., Kim, Y., Han, W., Herrasti, A., et al.: Imitating shortest paths in simulation enables effective navigation and manipulation in the real world. arXiv preprint arXiv:2312.02976 (2023) 6
2. Gadre, S.Y., Wortsman, M., Ilharco, G., Schmidt, L., Song, S.: CLIP on wheels: Zero-shot object navigation as object localization and exploration. CoRR **abs/2203.10421** (2022) 1
3. Krantz, J., Lee, S., Malik, J., Batra, D., Chaplot, D.S.: Instance-specific image goal navigation: Training embodied agents to find object instances. arXiv preprint arXiv:2211.15876 (2022) 1
4. Loper, E., Bird, S.: Nltk: The natural language toolkit. arXiv preprint cs/0205028 (2002) 4, 7
5. Majumdar, A., Aggarwal, G., Devnani, B., Hoffman, J., Batra, D.: ZSON: zero-shot object-goal navigation using multimodal goal embeddings. In: Proceedings of the International Conference on Neural Information Processing Systems (2022) 3, 7
6. Muhammad, M.B., Yeasin, M.: Eigen-cam: Class activation map using principal components. In: International Joint Conference on Neural Networks (IJCNN). pp. 1–7. IEEE (2020) 2
7. Radford, A., Kim, J.W., Hallacy, C., Ramesh, A., Goh, G., Agarwal, S., Sastry, G., Askell, A., Mishkin, P., Clark, J., Krueger, G., Sutskever, I.: Learning transferable visual models from natural language supervision. In: Meila, M., Zhang, T. (eds.) Proceedings of the International Conference on Machine Learning. vol. 139, pp. 8748–8763 (2021) 6
8. Rassin, R., Hirsch, E., Glickman, D., Ravfogel, S., Goldberg, Y., Chechik, G.: Linguistic binding in diffusion models: Enhancing attribute correspondence through attention map alignment. In: Proceedings of the International Conference on Neural Information Processing Systems. vol. 36 (2024) 4
9. Rombach, R., Blattmann, A., Lorenz, D., Esser, P., Ommer, B.: High-resolution image synthesis with latent diffusion models. In: Proceedings of the IEEE/CVF Conference on Computer Vision and Pattern Recognition. pp. 10684–10695 (2022) 4
10. Wang, W., Lv, Q., Yu, W., Hong, W., Qi, J., Wang, Y., Ji, J., Yang, Z., Zhao, L., Song, X., et al.: Cogvlm: Visual expert for pretrained language models. arXiv preprint arXiv:2311.03079 (2023) 1
11. Yadav, K., Ramrakhya, R., Majumdar, A., Berges, V., Kuhar, S., Batra, D., Baevski, A., Maksymets, O.: Offline visual representation learning for embodied navigation. CoRR **abs/2204.13226** (2022) 2
12. Zhai, X., Mustafa, B., Kolesnikov, A., Beyer, L.: Sigmoid loss for language image pre-training. In: Proceedings of the IEEE/CVF International Conference on Computer Vision. pp. 11975–11986 (2023) 6
13. Zhou, K., Zheng, K., Pryor, C., Shen, Y., Jin, H., Getoor, L., Wang, X.E.: ESC: exploration with soft commonsense constraints for zero-shot object navigation. In: Proceedings of the International Conference on Machine Learning. Proceedings of Machine Learning Research, vol. 202, pp. 42829–42842. PMLR (2023) 1



\end{document}

%% file: def.tex
\def\eg{\emph{e.g.,~}} 
\def\ie{\emph{i.e.,~}} 
 
\def\etc{\emph{etc.}} \def\vs{\emph{vs.~}}
\def\wrt{{w.r.t.~}}

\def\ba{{\bf a}}
\def\bb{{\bf b}}

\def\bh{{\bf h}}

\def\bp{{\bf p}}
\def\bq{{\bf q}}

\def\bs{{\bf s}}

\def\bv{{\bf v}}

\def\bz{{\bf z}}
\def\bI{{\bf I}}

\def\mC{{\mathcal C}}

\def\mV{{\mathcal V}}

\def\wrt{{w.r.t.~}}

\def\imagenav{$\texttt{ImageNav}$\xspace}
\def\objectnav{$\texttt{ObjectNav}$\xspace}
\def\instancenav{$\texttt{InstanceNav}$\xspace}

\def\sexyname{PSL\xspace}
\def\fullname{Prioritized Semantic Learning\xspace}
\def\lowername{prioritized semantic learning\xspace}

\def\spmname{SPM\xspace}
\def\spmfullname{Semantic Perception Module\xspace}
\def\spmlowername{semantic perception module\xspace}

\def\trainfullname{Prioritized Semantic Training\xspace}
\def\trainlowername{prioritized semantic training\xspace}

\def\inferfullname{Semantic Expansion Inference\xspace}
\def\inferlowername{semantic expansion inference\xspace}

\definecolor{mygray}{rgb}{0.9,0.9,0.9}
\definecolor{mygraydarker}{rgb}{0.7,0.7,0.7}
\definecolor{blackpink}{rgb}{0.83, 0.19, 0.79}
\definecolor{darkgreen}{rgb}{0.00, 0.74, 0.25}
\definecolor{blackgreen}{rgb}{0.00, 0.50, 0.10}
\def\notsure{\textcolor{black}}

\def\lec{\textcolor{black}}

\newcommand{\with}{\textcolor{darkgreen}{\ding{52}}}
\newcommand{\without}{\textcolor{red}{\ding{56}}}

%% file: arxiv.bbl
\begin{thebibliography}{10}
\providecommand{\url}[1]{\texttt{#1}}
\providecommand{\urlprefix}{URL }
\providecommand{\doi}[1]{https://doi.org/#1}

\bibitem{zer}
Al{-}Halah, Z., Ramakrishnan, S.K., Grauman, K.: Zero experience required: Plug {\&} play modular transfer learning for semantic visual navigation. In: Proceedings of the IEEE/CVF Conference on Computer Vision and Pattern Recognition. pp. 17010--17020 (2022)

\bibitem{anderson2018r2r}
Anderson, P., Wu, Q., Teney, D., Bruce, J., Johnson, M., S{\"u}nderhauf, N., Reid, I., Gould, S., Van Den~Hengel, A.: Vision-and-language navigation: Interpreting visually-grounded navigation instructions in real environments. In: Proceedings of the IEEE/CVF Conference on Computer Vision and Pattern Recognition. pp. 3674--3683 (2018)

\bibitem{R2R}
Anderson, P., Wu, Q., Teney, D., Bruce, J., Johnson, M., S{\"{u}}nderhauf, N., Reid, I.D., Gould, S., van~den Hengel, A.: Vision-and-language navigation: Interpreting visually-grounded navigation instructions in real environments. In: Proceedings of the IEEE/CVF Conference on Computer Vision and Pattern Recognition (2018)

\bibitem{Batra2020rearrange}
Batra, D., Chang, A.X., Chernova, S., Davison, A.J., Deng, J., Koltun, V., Levine, S., Malik, J., Mordatch, I., Mottaghi, R., Savva, M., Su, H.: Rearrangement: {A} challenge for embodied {AI}. CoRR  \textbf{abs/2011.01975} (2020)

\bibitem{batra2020objectnav}
Batra, D., Gokaslan, A., Kembhavi, A., Maksymets, O., Mottaghi, R., Savva, M., Toshev, A., Wijmans, E.: Objectnav revisited: On evaluation of embodied agents navigating to objects. arXiv preprint arXiv:2006.13171  (2020)

\bibitem{pixelnav}
Cai, W., Huang, S., Cheng, G., Long, Y., Gao, P., Sun, C., Dong, H.: Bridging zero-shot object navigation and foundation models through pixel-guided navigation skill. CoRR  \textbf{abs/2309.10309} (2023)

\bibitem{mp3d}
Chang, A., Dai, A., Funkhouser, T., Halber, M., Niessner, M., Savva, M., Song, S., Zeng, A., Zhang, Y.: Matterport3d: Learning from rgb-d data in indoor environments. arXiv preprint arXiv:1709.06158  (2017)

\bibitem{a2nav}
Chen, P., Sun, X., Zhi, H., Zeng, R., Li, T.H., Liu, G., Tan, M., Gan, C.: $a^2$ nav: Action-aware zero-shot robot navigation by exploiting vision-and-language ability of foundation models. arXiv preprint arXiv:2308.07997  (2023)

\bibitem{procthor}
Deitke, M., VanderBilt, E., Herrasti, A., Weihs, L., Ehsani, K., Salvador, J., Han, W., Kolve, E., Kembhavi, A., Mottaghi, R.: Procthor: Large-scale embodied {AI} using procedural generation. In: Koyejo, S., Mohamed, S., Agarwal, A., Belgrave, D., Cho, K., Oh, A. (eds.) Proceedings of the International Conference on Neural Information Processing Systems (2022)

\bibitem{crl}
Du, Y., Gan, C., Isola, P.: Curious representation learning for embodied intelligence. In: Proceedings of the IEEE/CVF International Conference on Computer Vision (2021)

\bibitem{cow}
Gadre, S.Y., Wortsman, M., Ilharco, G., Schmidt, L., Song, S.: {CLIP} on wheels: Zero-shot object navigation as object localization and exploration. CoRR  \textbf{abs/2203.10421} (2022)

\bibitem{gan2022finding}
Gan, C., Gu, Y., Zhou, S., Schwartz, J., Alter, S., Traer, J., Gutfreund, D., Tenenbaum, J.B., McDermott, J.H., Torralba, A.: Finding fallen objects via asynchronous audio-visual integration. In: Proceedings of the IEEE/CVF Conference on Computer Vision and Pattern Recognition. pp. 10523--10533 (2022)

\bibitem{gan2020threedworld}
Gan, C., Schwartz, J., Alter, S., Mrowca, D., Schrimpf, M., Traer, J., De~Freitas, J., Kubilius, J., Bhandwaldar, A., Haber, N., et~al.: Threedworld: A platform for interactive multi-modal physical simulation. arXiv preprint arXiv:2007.04954  (2020)

\bibitem{gao2023llama}
Gao, P., Han, J., Zhang, R., Lin, Z., Geng, S., Zhou, A., Zhang, W., Lu, P., He, C., Yue, X., et~al.: Llama-adapter v2: Parameter-efficient visual instruction model. arXiv preprint arXiv:2304.15010  (2023)

\bibitem{nrns}
Hahn, M., Chaplot, D.S., Tulsiani, S., Mukadam, M., Rehg, J.M., Gupta, A.: No rl, no simulation: Learning to navigate without navigating. In: Proceedings of the International Conference on Neural Information Processing Systems (2021)

\bibitem{he2016deep}
He, K., Zhang, X., Ren, S., Sun, J.: Deep residual learning for image recognition. In: Proceedings of the IEEE/CVF Conference on Computer Vision and Pattern Recognition. pp. 770--778 (2016)

\bibitem{jia2021scaling}
Jia, C., Yang, Y., Xia, Y., Chen, Y.T., Parekh, Z., Pham, H., Le, Q., Sung, Y.H., Li, Z., Duerig, T.: Scaling up visual and vision-language representation learning with noisy text supervision. In: Proceedings of the International Conference on Machine Learning. pp. 4904--4916. PMLR (2021)

\bibitem{tsgm}
Kim, N., Kwon, O., Yoo, H., Choi, Y., Park, J., Oh, S.: Topological semantic graph memory for image-goal navigation. In: Conference on Robot Learning. pp. 393--402. PMLR (2023)

\bibitem{modiin}
Krantz, J., Gervet, T., Yadav, K., Wang, A., Paxton, C., Mottaghi, R., Batra, D., Malik, J., Lee, S., Chaplot, D.S.: Navigating to objects specified by images. In: Proceedings of the IEEE/CVF International Conference on Computer Vision. pp. 10916--10925 (2023)

\bibitem{krantz2021waypoint}
Krantz, J., Gokaslan, A., Batra, D., Lee, S., Maksymets, O.: Waypoint models for instruction-guided navigation in continuous environments. In: Proceedings of the IEEE/CVF International Conference on Computer Vision. pp. 15162--15171 (2021)

\bibitem{iin}
Krantz, J., Lee, S., Malik, J., Batra, D., Chaplot, D.S.: Instance-specific image goal navigation: Training embodied agents to find object instances. arXiv preprint arXiv:2211.15876  (2022)

\bibitem{krantz2020vlnce}
Krantz, J., Wijmans, E., Majumdar, A., Batra, D., Lee, S.: Beyond the nav-graph: Vision-and-language navigation in continuous environments. In: European Conference on Computer Vision. pp. 104--120. Springer (2020)

\bibitem{RxR}
Ku, A., Anderson, P., Patel, R., Ie, E., Baldridge, J.: {Room-Across-Room}: Multilingual vision-and-language navigation with dense spatiotemporal grounding. In: Conference on Empirical Methods for Natural Language Processing (2020)

\bibitem{vgm}
Kwon, O., Kim, N., Choi, Y., Yoo, H., Park, J., Oh, S.: Visual graph memory with unsupervised representation for visual navigation. In: Proceedings of the IEEE/CVF International Conference on Computer Vision. pp. 15890--15899 (2021)

\bibitem{li2021igibson}
Li, C., Xia, F., Mart{\'\i}n-Mart{\'\i}n, R., Lingelbach, M., Srivastava, S., Shen, B., Vainio, K., Gokmen, C., Dharan, G., Jain, T., et~al.: igibson 2.0: Object-centric simulation for robot learning of everyday household tasks. arXiv preprint arXiv:2108.03272  (2021)

\bibitem{iGibson}
Li, C., Xia, F., Mart{\'{\i}}n{-}Mart{\'{\i}}n, R., Lingelbach, M., Srivastava, S., Shen, B., Vainio, K.E., Gokmen, C., Dharan, G., Jain, T., Kurenkov, A., Liu, C.K., Gweon, H., Wu, J., Fei{-}Fei, L., Savarese, S.: igibson 2.0: Object-centric simulation for robot learning of everyday household tasks. In: Conference on Robot Learning. vol.~164, pp. 455--465 (2021)

\bibitem{BEHAVIOR-1K}
Li, C., Zhang, R., Wong, J., Gokmen, C., Srivastava, S., Mart{\'{\i}}n{-}Mart{\'{\i}}n, R., Wang, C., Levine, G., Lingelbach, M., Sun, J., Anvari, M., Hwang, M., Sharma, M., Aydin, A., Bansal, D., Hunter, S., Kim, K., Lou, A., Matthews, C.R., Villa{-}Renteria, I., Tang, J.H., Tang, C., Xia, F., Savarese, S., Gweon, H., Liu, K., Wu, J., Fei{-}Fei, L.: {BEHAVIOR-1K:} {A} benchmark for embodied {AI} with 1, 000 everyday activities and realistic simulation. In: Conference on Robot Learning. vol.~205, pp. 80--93 (2022)

\bibitem{li2023behavior}
Li, C., Zhang, R., Wong, J., Gokmen, C., Srivastava, S., Mart{\'\i}n-Mart{\'\i}n, R., Wang, C., Levine, G., Lingelbach, M., Sun, J., et~al.: Behavior-1k: A benchmark for embodied ai with 1,000 everyday activities and realistic simulation. In: Conference on Robot Learning. pp. 80--93. PMLR (2023)

\bibitem{li2023blip}
Li, J., Li, D., Savarese, S., Hoi, S.: Blip-2: Bootstrapping language-image pre-training with frozen image encoders and large language models. arXiv preprint arXiv:2301.12597  (2023)

\bibitem{li2021align}
Li, J., Selvaraju, R., Gotmare, A., Joty, S., Xiong, C., Hoi, S.C.H.: Align before fuse: Vision and language representation learning with momentum distillation. Proceedings of the International Conference on Neural Information Processing Systems  \textbf{34},  9694--9705 (2021)

\bibitem{li2019visualbert}
Li, L.H., Yatskar, M., Yin, D., Hsieh, C.J., Chang, K.W.: Visualbert: A simple and performant baseline for vision and language. arXiv preprint arXiv:1908.03557  (2019)

\bibitem{li2023decap}
Li, W., Zhu, L., Wen, L., Yang, Y.: Decap: Decoding clip latents for zero-shot captioning via text-only training. arXiv preprint arXiv:2303.03032  (2023)

\bibitem{liang2022mind}
Liang, V.W., Zhang, Y., Kwon, Y., Yeung, S., Zou, J.Y.: Mind the gap: Understanding the modality gap in multi-modal contrastive representation learning. In: Proceedings of the International Conference on Neural Information Processing Systems. vol.~35, pp. 17612--17625 (2022)

\bibitem{liu2023improved}
Liu, H., Li, C., Li, Y., Lee, Y.J.: Improved baselines with visual instruction tuning. arXiv preprint arXiv:2310.03744  (2023)

\bibitem{liu2023llava}
Liu, H., Li, C., Wu, Q., Lee, Y.J.: Visual instruction tuning. arXiv preprint arXiv:2304.08485  (2023)

\bibitem{llava}
Liu, H., Li, C., Wu, Q., Lee, Y.J.: Visual instruction tuning. arXiv preprint arXiv:2304.08485  (2023)

\bibitem{ma2023simple}
Ma, S., Wang, Y., Wei, Y., Fan, J., Sun, X., Chen, P., Zhang, E.: A simple knowledge distillation framework for open-world object detection. arXiv preprint arXiv:2312.08653  (2023)

\bibitem{zson}
Majumdar, A., Aggarwal, G., Devnani, B., Hoffman, J., Batra, D.: {ZSON:} zero-shot object-goal navigation using multimodal goal embeddings. In: Proceedings of the International Conference on Neural Information Processing Systems (2022)

\bibitem{majumdar2023findthis}
Majumdar, A., Xia, F., Batra, D., Guibas, L., et~al.: Findthis: Language-driven object disambiguation in indoor environments. In: Conference on Robot Learning (2023)

\bibitem{mem-aug}
Mezghani, L., Sukhbaatar, S., Lavril, T., Maksymets, O., Batra, D., Bojanowski, P., Alahari, K.: Memory-augmented reinforcement learning for image-goal navigation. In: IEEE/RSJ International Conference on Intelligent Robots and Systems (2022)

\bibitem{gpt4}
OpenAI: {GPT-4} technical report. CoRR  \textbf{abs/2303.08774} (2023)

\bibitem{clip}
Radford, A., Kim, J.W., Hallacy, C., Ramesh, A., Goh, G., Agarwal, S., Sastry, G., Askell, A., Mishkin, P., Clark, J., Krueger, G., Sutskever, I.: Learning transferable visual models from natural language supervision. In: Meila, M., Zhang, T. (eds.) Proceedings of the International Conference on Machine Learning. vol.~139, pp. 8748--8763 (2021)

\bibitem{hm3d}
Ramakrishnan, S.K., Gokaslan, A., Wijmans, E., Maksymets, O., Clegg, A., Turner, J., Undersander, E., Galuba, W., Westbury, A., Chang, A.X., et~al.: Habitat-matterport 3d dataset (hm3d): 1000 large-scale 3d environments for embodied ai. arXiv preprint arXiv:2109.08238  (2021)

\bibitem{ramrakhya2023pirlnav}
Ramrakhya, R., Batra, D., Wijmans, E., Das, A.: Pirlnav: Pretraining with imitation and rl finetuning for objectnav. In: Proceedings of the IEEE/CVF Conference on Computer Vision and Pattern Recognition. pp. 17896--17906 (2023)

\bibitem{habitat}
Savva, M., Malik, J., Parikh, D., Batra, D., Kadian, A., Maksymets, O., Zhao, Y., Wijmans, E., Jain, B., Straub, J., Liu, J., Koltun, V.: Habitat: {A} platform for embodied {AI} research. In: Proceedings of the IEEE/CVF International Conference on Computer Vision (2019)

\bibitem{PPO}
Schulman, J., Wolski, F., Dhariwal, P., Radford, A., Klimov, O.: Proximal policy optimization algorithms. arXiv preprint arXiv:1707.06347  (2017)

\bibitem{BEHAVIOR}
Srivastava, S., Li, C., Lingelbach, M., Mart{\'{\i}}n{-}Mart{\'{\i}}n, R., Xia, F., Vainio, K.E., Lian, Z., Gokmen, C., Buch, S., Liu, C.K., Savarese, S., Gweon, H., Wu, J., Fei{-}Fei, L.: {BEHAVIOR:} benchmark for everyday household activities in virtual, interactive, and ecological environments. In: Conference on Robot Learning. vol.~164, pp. 477--490 (2021)

\bibitem{srivastava2022behavior}
Srivastava, S., Li, C., Lingelbach, M., Mart{\'\i}n-Mart{\'\i}n, R., Xia, F., Vainio, K.E., Lian, Z., Gokmen, C., Buch, S., Liu, K., et~al.: Behavior: Benchmark for everyday household activities in virtual, interactive, and ecological environments. In: Conference on Robot Learning. pp. 477--490. PMLR (2022)

\bibitem{habitat2}
Szot, A., Clegg, A., Undersander, E., Wijmans, E., Zhao, Y., Turner, J., Maestre, N., Mukadam, M., Chaplot, D.S., Maksymets, O., Gokaslan, A., Vondrus, V., Dharur, S., Meier, F., Galuba, W., Chang, A.X., Kira, Z., Koltun, V., Malik, J., Savva, M., Batra, D.: Habitat 2.0: Training home assistants to rearrange their habitat. In: Proceedings of the International Conference on Neural Information Processing Systems (2021)

\bibitem{thomason2020dialog}
Thomason, J., Murray, M., Cakmak, M., Zettlemoyer, L.: Vision-and-dialog navigation. In: Conference on Robot Learning. pp. 394--406. PMLR (2020)

\bibitem{udandarao2023visual}
Udandarao, V., Burg, M.F., Albanie, S., Bethge, M.: Visual data-type understanding does not emerge from scaling vision-language models. In: Proceedings of the International Conference on Learning Representations (2023)

\bibitem{cogvlm}
Wang, W., Lv, Q., Yu, W., Hong, W., Qi, J., Wang, Y., Ji, J., Yang, Z., Zhao, L., Song, X., et~al.: Cogvlm: Visual expert for pretrained language models. arXiv preprint arXiv:2311.03079  (2023)

\bibitem{ai2thorRearrange}
Weihs, L., Deitke, M., Kembhavi, A., Mottaghi, R.: Visual room rearrangement. In: Proceedings of the IEEE/CVF Conference on Computer Vision and Pattern Recognition. pp. 5922--5931 (2021)

\bibitem{DDPPO}
Wijmans, E., Kadian, A., Morcos, A., Lee, S., Essa, I., Parikh, D., Savva, M., Batra, D.: {DD-PPO:} learning near-perfect pointgoal navigators from 2.5 billion frames. In: Proceedings of the International Conference on Learning Representations (2020)

\bibitem{wu2018roomnav}
Wu, Y., Wu, Y., Gkioxari, G., Tian, Y.: Building generalizable agents with a realistic and rich 3d environment. arXiv preprint arXiv:1801.02209  (2018)

\bibitem{fgprompt}
Xinyu, S., Peihao, C., Jugang, F., Thomas, H.L., Jian, C., Mingkui, T.: Fgprompt: Fine-grained goal prompting for image-goal navigation. In: Proceedings of the International Conference on Neural Information Processing Systems (2023)

\bibitem{ovrlv2}
Yadav, K., Majumdar, A., Ramrakhya, R., Yokoyama, N., Baevski, A., Kira, Z., Maksymets, O., Batra, D.: Ovrl-v2: A simple state-of-art baseline for imagenav and objectnav. arXiv preprint arXiv:2303.07798  (2023)

\bibitem{ovrl}
Yadav, K., Ramrakhya, R., Majumdar, A., Berges, V., Kuhar, S., Batra, D., Baevski, A., Maksymets, O.: Offline visual representation learning for embodied navigation. CoRR  \textbf{abs/2204.13226} (2022)

\bibitem{hm3dv2}
Yadav, K., Ramrakhya, R., Ramakrishnan, S.K., Gervet, T., Turner, J., Gokaslan, A., Maestre, N., Chang, A.X., Batra, D., Savva, M., et~al.: Habitat-matterport 3d semantics dataset. In: Proceedings of the IEEE/CVF Conference on Computer Vision and Pattern Recognition. pp. 4927--4936 (2023)

\bibitem{yamauchi1997frontier}
Yamauchi, B.: A frontier-based approach for autonomous exploration. In: Proceedings 1997 IEEE International Symposium on Computational Intelligence in Robotics and Automation CIRA'97. pp. 146--151. IEEE (1997)

\bibitem{yenamandra2023homerobot}
Yenamandra, S., Ramachandran, A., Yadav, K., Wang, A., Khanna, M., Gervet, T., Yang, T.Y., Jain, V., Clegg, A.W., Turner, J., et~al.: Homerobot: Open-vocabulary mobile manipulation. arXiv preprint arXiv:2306.11565  (2023)

\bibitem{yu2023l3mvn}
Yu, B., Kasaei, H., Cao, M.: L3mvn: Leveraging large language models for visual target navigation. arXiv preprint arXiv:2304.05501  (2023)

\bibitem{zhang2023llama}
Zhang, R., Han, J., Zhou, A., Hu, X., Yan, S., Lu, P., Li, H., Gao, P., Qiao, Y.: Llama-adapter: Efficient fine-tuning of language models with zero-init attention. arXiv preprint arXiv:2303.16199  (2023)

\bibitem{esc}
Zhou, K., Zheng, K., Pryor, C., Shen, Y., Jin, H., Getoor, L., Wang, X.E.: {ESC:} exploration with soft commonsense constraints for zero-shot object navigation. In: Proceedings of the International Conference on Machine Learning. Proceedings of Machine Learning Research, vol.~202, pp. 42829--42842. {PMLR} (2023)

\bibitem{zhu2021soon}
Zhu, F., Liang, X., Zhu, Y., Yu, Q., Chang, X., Liang, X.: Soon: Scenario oriented object navigation with graph-based exploration. In: Proceedings of the IEEE/CVF Conference on Computer Vision and Pattern Recognition. pp. 12689--12699 (2021)

\bibitem{imagenav}
Zhu, Y., Mottaghi, R., Kolve, E., Lim, J.J., Gupta, A., Fei{-}Fei, L., Farhadi, A.: Target-driven visual navigation in indoor scenes using deep reinforcement learning. In: IEEE International Conference on Robotics and Automation (2017)

\end{thebibliography}
